\documentclass[journal]{IEEEtran}

\usepackage{cite}
\usepackage{amsmath,amssymb,amsfonts}
\usepackage{graphicx}
\usepackage{textcomp}
\usepackage{xcolor}
\usepackage{booktabs}
\usepackage{multirow}
\usepackage{diagbox}
\usepackage{hyperref}
\usepackage{array}
\usepackage{url}
\usepackage{bm}
\usepackage{balance}
\usepackage{algorithm}
\usepackage{algorithmic}
\usepackage[caption=false,font=footnotesize]{subfig}
\usepackage{float}

\hypersetup{
 colorlinks=true,
 linkcolor=blue,
 citecolor=blue,
 urlcolor=blue,
}

\begin{document}
\bstctlcite{IEEEexample:BSTcontrol}

\title{Geometry-Aware Cross-Height Channel Knowledge Map Prediction for UAV-Assisted Communications With Uncertainty-Guided 3D Sensing}

\author{Zhihan Zeng, Amir Hussain, \IEEEmembership{Senior Member, IEEE}, Yue Xiu, \IEEEmembership{Member, IEEE}, \\Phee Lep Yeoh, \IEEEmembership{Senior Member, IEEE},
Lu Chen, Zhongpei Zhang, Guan Gui, \IEEEmembership{Fellow, IEEE}

\thanks{Zhihan Zeng, Yue Xiu and Zhongpei Zhang are with the National Key Laboratory of Wireless Communications, University of Electronic Science and Technology of China (UESTC), Chengdu 611731, China (E-mail: 202511220608@std.uestc.edu.cn, xiuyue12345678@163.com, zhangzp@uestc.edu.cn). Amir Hussain is with the SDAIA-KFUPM Joint Research Centre of AI, King Fahd University of Petroleum and Minerals, Dhahran, Saudi Arabia, (E-mail: amir.hussain@kfupm.edu.sa). Phee Lep Yeoh is with the School of Science, Technology and Engineering, University of the Sunshine Coast, Moreton Bay Campus, Petrie, QLD 4502, Australia (E-mail:pyeoh@usc.edu.au). Lu Chen is with the Anhui Science and Technology University Chuzhou 239000, China and Nanjing University Of Information Science $\&$ technology, Nanjing 210044, China (E-mail: chenlu@ahstu.edu.cn). Guan Gui is with the College of Telecommunications and Information Engineering, Nanjing University of Posts and Telecommunications, Nanjing 210003, China (E-mail:guiguan@njupt.edu.cn). The corresponding author is Lu Chen.}
}

\maketitle

\begin{abstract}
Low-altitude Unmanned Aerial Vehicles (UAVs) often need to infer channel knowledge across a range of heights from only sparse observations collected at a few altitude layers. To address this challenge, this paper studies height-conditioned cross-height channel knowledge map (CKM) prediction for UAV-assisted communications in geometry-rich urban environments. We develop a geometry-aware conditional prediction framework that combines urban scene priors, sparse multi-altitude observations, and target-height descriptors to reconstruct dense CKMs at unobserved target heights. An uncertainty head is further introduced to characterize prediction confidence and to support cost-aware online UAV sensing under motion and safety constraints. Experiments on a layered aerial CKM benchmark show that the proposed Feature Pyramid Network (FPN)-Transformer achieves the best overall performance under both unseen-scene zero-shot and legacy patch-random protocols, reducing the Root Mean Square Error (RMSE) to 5.347~dB and 1.111~dB, respectively, compared with 6.937~dB and 1.221~dB for the strongest baseline 3D-RadioDiff. Moreover, after applying our unseen-scene few-shot adaptation, the RMSE further decreases from 5.347~dB in zero-shot prediction to 3.518~dB with 10-shot two-height support, while the uncertainty-guided cost-aware sensing policy improves active reconstruction from 6.94~dB at initialization to 4.79~dB at sensing budget 40, outperforming uncertainty-only sensing at 5.08~dB and random aerial sampling at 5.84~dB.
\end{abstract}


\section{Introduction}
\label{sec_intro}

\IEEEPARstart{A}{s} wireless systems evolve toward IMT-2030 and future sixth-generation (6G) networks, communication infrastructures are expected to move beyond conventional connectivity provisioning toward more intelligent, environment-aware, and service-adaptive operation. In this vision, low-altitude intelligent networking, ubiquitous connectivity, integrated sensing and communication, and AI-native wireless intelligence are becoming increasingly important for next-generation mobile systems. Within this broader trend, low-altitude unmanned aerial vehicles (UAVs) are expected to play an increasingly important role in future intelligent wireless systems for emergency response, temporary coverage restoration, rapid environment exploration, and on-demand communication support \cite{ITURM2160,Sun2026KnowledgeDrivenDL6G,Jiang2026UniRM,Nguyen2022Survey6GIoT,Saad2020Vision6G,Kato2019AISAGIN,zeng2026urbanrtrm,Liu2018SAGINSurvey}. In parallel, environment-aware learning is also attracting attention in beyond-terrestrial and special-environment wireless systems, including deep-space and satellite-assisted settings \cite{Cai2025DeepSpaceChannelEstimation,Gao2025LEODeepSpacePerformance,zeng2026jsrgfnet,zeng2026skanet}.

In such scenarios, a UAV often needs to quickly understand the propagation environment of an unfamiliar urban area before reliable communication, relay deployment, or environment-aware networking decisions can be made. This requirement makes CKM estimation a key enabler for UAV-assisted communications and for broader intelligent wireless operation in emerging 6G-style environments. However, the challenge in low-altitude aerial communication is not merely sparse spatial sampling on a horizontal plane. In geometry-rich urban environments, the propagation field changes significantly with altitude because line-of-sight conditions, blockage transitions, diffraction behavior, and geometry-dependent multipath all evolve as the UAV moves vertically. As a result, observations collected at one height cannot be directly reused for another height, while exhaustive measurement over the full three-dimensional aerial space is impractical under strict motion, time, and flight-safety constraints. Therefore, the practical challenge for UAV-assisted communications is to infer channel conditions at unmeasured target heights from only sparse observations collected at a few accessible altitude layers \cite{wang2025consistency,Romero2018BlindRadioTomography,ai2025openset}.

Existing terrestrial CKM and UAV-related aerial CKM construction studies have established the feasibility of learning environment-aware propagation fields from sparse observations \cite{Wu2018AutomaticRadioMapAdaptation,zeng2026phygmoe,Sato2021SpaceFrequencyRadioMap,Chen2025DynamicSpectrumCartography,zeng2026gackan}. However, most existing methods are developed for terrestrial or single-height reconstruction, same-context CKM inference, fixed-service-layer aerial CKM construction, or downstream task optimization. They do not directly formulate height-conditioned cross-height CKM prediction from sparse multi-altitude observations as the primary learning problem.

Motivated by emergency deployment and rapid aerial exploration, this paper studies height-conditioned cross-height CKM prediction for UAV-assisted communications. Specifically, given sparse observations collected at several observed heights, together with scene geometry and target-height descriptors, the objective is to reconstruct the dense CKM at an unobserved target height. Under this task formulation, uncertainty estimation, active sensing, and few-shot adaptation are not treated as co-equal primary tasks. Instead, they are introduced as deployment-oriented extensions for reliability assessment, online sensing utility, and rapid adaptation in unseen scenes.

To solve this problem, we develop a geometry-aware conditional prediction framework built on an FPN-Transformer backbone. The model fuses explicit urban scene priors, sparse height-indexed observations, and target-height descriptors to reconstruct dense target-height CKMs at unmeasured altitudes. On top of the primary predictor, we further introduce an auxiliary uncertainty head and evaluate its usefulness in cost-aware sensing under aerial motion and safety constraints. We also study few-shot adaptation to assess rapid deployment capability in previously unseen scenes.

Compared with conventional single-height CKM reconstruction, the present work shifts the focus from plane-wise completion to layered cross-height conditional inference. It also changes the sensing setting from unconstrained or accessibility-limited two-dimensional sampling to informative aerial measurement selection in a flyable three-dimensional space with motion cost and risk awareness. This task shift is important because it better matches how UAVs are actually deployed in unfamiliar urban environments and how low-altitude intelligent networking is expected to operate in IMT-2030-oriented wireless systems.

The main contributions of this paper are summarized as follows.
\begin{itemize}
 \item We formulate the problem of height-conditioned cross-height CKM prediction for UAV-assisted communications, where dense target-height CKM slices are inferred from sparse observations collected at a few observed altitude layers.

 \item We develop a geometry-aware conditional prediction framework that combines urban scene priors, sparse multi-altitude observations, and target-height descriptors for dense CKM reconstruction at unmeasured heights.

 \item We construct a layered aerial CKM benchmark and evaluate the proposed framework under unseen-scene, cross-height, and rapid-deployment settings, including zero-shot and few-shot protocols.

 \item We further show that the predicted uncertainty can support cost-aware online sensing under limited UAV motion budgets, serving as a system-level extension of the primary prediction framework.
\end{itemize}

The remainder of this paper is organized as follows. Section~\ref{sec_related_work} reviews the related work. Section~\ref{sec_system_model} defines the scene model, aerial constraints, and the primary cross-height prediction task. Section~\ref{sec_method} presents the proposed geometry-aware prediction framework together with the downstream uncertainty-guided sensing strategy. Section~\ref{sec_experiments} describes the dataset construction and experimental setup. Section~\ref{sec_results} reports the numerical results and analysis. Finally, Section~\ref{sec_conclusion} concludes this paper.

\section{Related Work}
\label{sec_related_work}

\subsection{Terrestrial and Single-Height CKM Reconstruction}
Early location-indexed channel-environment estimation studies mainly focused on spectrum cartography, radio tomography, and coverage or CKM reconstruction using model-based, kernel-based, and low-rank or sparse formulations. Representative works include distributed spectrum sensing and map recovery \cite{Bazerque2010DistributedSpectrumSensingSparsity}, kernel-based adaptive coverage map reconstruction \cite{Kasparick2016KernelOnlineCoverageMap}, channel gain cartography with low-rank and sparsity priors \cite{Lee2017ChannelGainCartographyLowRankSparse}, learning spectrum maps from quantized measurements \cite{Romero2017PSDMapQuantizedMeasurements}, blind radio tomography \cite{Romero2018BlindRadioTomography}, variance-based radio tomography \cite{Wilson2011VarianceBasedRTI}, accuracy-versus-resolution analysis \cite{Martin2014AccuracyVsResolutionRTI}, adaptive Bayesian radio tomography \cite{Lee2019AdaptiveBayesianRadioTomography}, location-free spectrum cartography \cite{Teganya2019LocationFreeSpectrumCartography}, automatic CKM adaptation \cite{Wu2018AutomaticRadioMapAdaptation}, and theoretical analysis of channel-map estimation \cite{Romero2024TheoreticalAnalysisRME}. These studies established the mathematical and statistical foundations of CKM reconstruction, but they were not designed for height-conditioned aerial inference from sparse multi-altitude observations.

More recently, deep learning has substantially improved dense CKM recovery from sparse measurements. Representative directions include convolutional learning \cite{Levie2021RadioUNet}, deep completion autoencoders \cite{Teganya2022DeepCompletionAutoencoders}, tensor-style neural completion \cite{Shrestha2022DeepSpectrumCartography}, deep reconstruction for V2X settings \cite{Roger2024DLRadioMapV2X}, graph-attention-based reconstruction \cite{Li2024RadioGAT}, restricted-area CKM inpainting \cite{Zhang2024RadiomapInpaintingRestrictedAreas}, deformable-attention CKM estimation \cite{Liu2026DATUnetRadioMap}, diffusion-based CKM construction \cite{Wang2025RadioDiff}, dynamic spectrum cartography \cite{Chen2025DynamicSpectrumCartography}, space-frequency interpolation \cite{Sato2021SpaceFrequencyRadioMap}, interpretable symbolic-data-fusion prediction, and KAN-inspired representation learning \cite{Liu2025KAN}, together with transfer-learning-based estimation. Active data acquisition has also been explored, for example through Bayesian active learning for sample-efficient CKM reconstruction \cite{Polyzos2024BayesianActiveLearningRadioMap}. Although these methods significantly advance sparse-to-dense reconstruction, most of them still operate in terrestrial, single-height, or same-distribution settings, where the queried CKM slice is reconstructed on the same spatial layer as the main sensing context. Thus, their limitation is not simply that they require dense observations; rather, they generally require target-layer observations or a same-context sensing setting, and do not explicitly infer a dense CKM slice at an unobserved altitude from sparse measurements collected at other heights.

\subsection{CKM and Cross-Context Prediction}
The CKM literature has expanded location-indexed channel representation ideas toward environment-aware channel prediction and communication optimization. Recent work has studied how much data is needed for CKM construction \cite{Xu2024HowMuchDataCKM}, CKM-aided channel prediction with measurement-based evaluation \cite{Wang2025CKMAidedChannelPrediction}, and CKM-assisted beamforming or communication optimization \cite{Wu2024EnvironmentAwareHybridBeamformingCKM,Yang2025RadioMapBeamformingReducedPilots}. These studies show that environment-aware channel representations can effectively support downstream wireless tasks.

Several recent works have further explored CKM generation under incomplete or shifted contexts. Examples include image-to-image style CKM inpainting \cite{Jin2025I2IInpaintingCKM}, cross-AP CKM inference using others' data \cite{Dai2025CrossAPCKM}, and diffusion-based CKM generation from partial observations \cite{zhao2025radiodiff}. In addition, CKM-style ideas have also been considered in low-altitude applications, such as CKM-assisted routing for the low-altitude economy and AIGC-oriented CKM construction for low-altitude channel estimation. Nevertheless, the main emphasis of these studies is still CKM construction, same-context channel inference, or cross-source transfer within related communication settings. Most existing CKM formulations are tied to a single altitude layer, a fixed service plane, or a same-context sensing domain, and extending them to multiple altitudes is nontrivial because the model must exploit vertical correlation while accounting for height-dependent line-of-sight transitions, blockage changes, and multipath variations. Thus, the communication settings at different altitudes are not unrelated in our task; rather, they are physically coupled through the same scene geometry and base-station deployment, but cannot be treated as directly interchangeable layers. They do not explicitly address the task considered here, namely dense target-height CKM prediction at an unobserved altitude from sparse observations collected across several other altitude layers under aerial flight constraints.

\subsection{UAV, 3D Aerial CKM Construction, and Sensing}
A growing body of work has considered the importance of three-dimensional (3D) CKMs in UAV and aerial communication scenarios. Simultaneous navigation and CKM construction for cellular-connected UAVs was studied in \cite{Zeng2021SNARMUAV}, while active CKM estimation with autonomous UAVs was investigated in \cite{Shrestha2023SpectrumSurveying}. Related aerial uses of CKMs include aerial base-station placement via propagation maps \cite{Romero2024AerialBaseStationPlacementRadioMaps}, CKM-assisted predictive UAV communications \cite{Li2024RadioMapPredictiveUAV}, and CKM-guided motion planning ideas that were also explored in robotic navigation settings \cite{Mu2021IRSRobotPathPlanningRadioMap}. These works clearly demonstrate the value of channel environment awareness in aerial systems.

More recent studies have moved toward explicitly aerial or 3D channel environments. Representative examples include CKM reconstruction for UAV communications \cite{Zhao2025RMCDDRUAV}, efficient urban 3D CKM estimation from sparse measurements \cite{Chen2025Urban3DRadioMapSparse}, and the SpectrumNet benchmark for multiband 3D CKMs. Although these efforts make important progress toward aerial and 3D propagation modeling, their formulations usually emphasize fixed-layer mapping, full-volume generation, benchmark construction, or downstream deployment tasks. They do not directly focus on the problem most relevant to rapid UAV deployment in unfamiliar urban environments: using sparse multi-altitude observations to infer the dense CKM at a queried, previously unmeasured target height.

Overall, the existing literature provides strong foundations for terrestrial CKM reconstruction, CKM-based environment-aware prediction, and UAV or three-dimensional aerial CKM construction. Nevertheless, these directions have not yet been fully unified into a single framework that directly addresses dense CKM prediction at an unmeasured target height from sparse multi-altitude observations, with explicit target-height conditioning and deployment-oriented sensing capability under aerial motion and safety constraints. This paper aims to fill this gap by treating cross-height aerial CKM prediction as the primary learning problem and uncertainty-guided online sensing as a downstream extension built upon that predictor.

\section{System Model}
\label{sec_system_model}

\begin{figure}[htpb]
 \centering
 \includegraphics[width=\columnwidth]{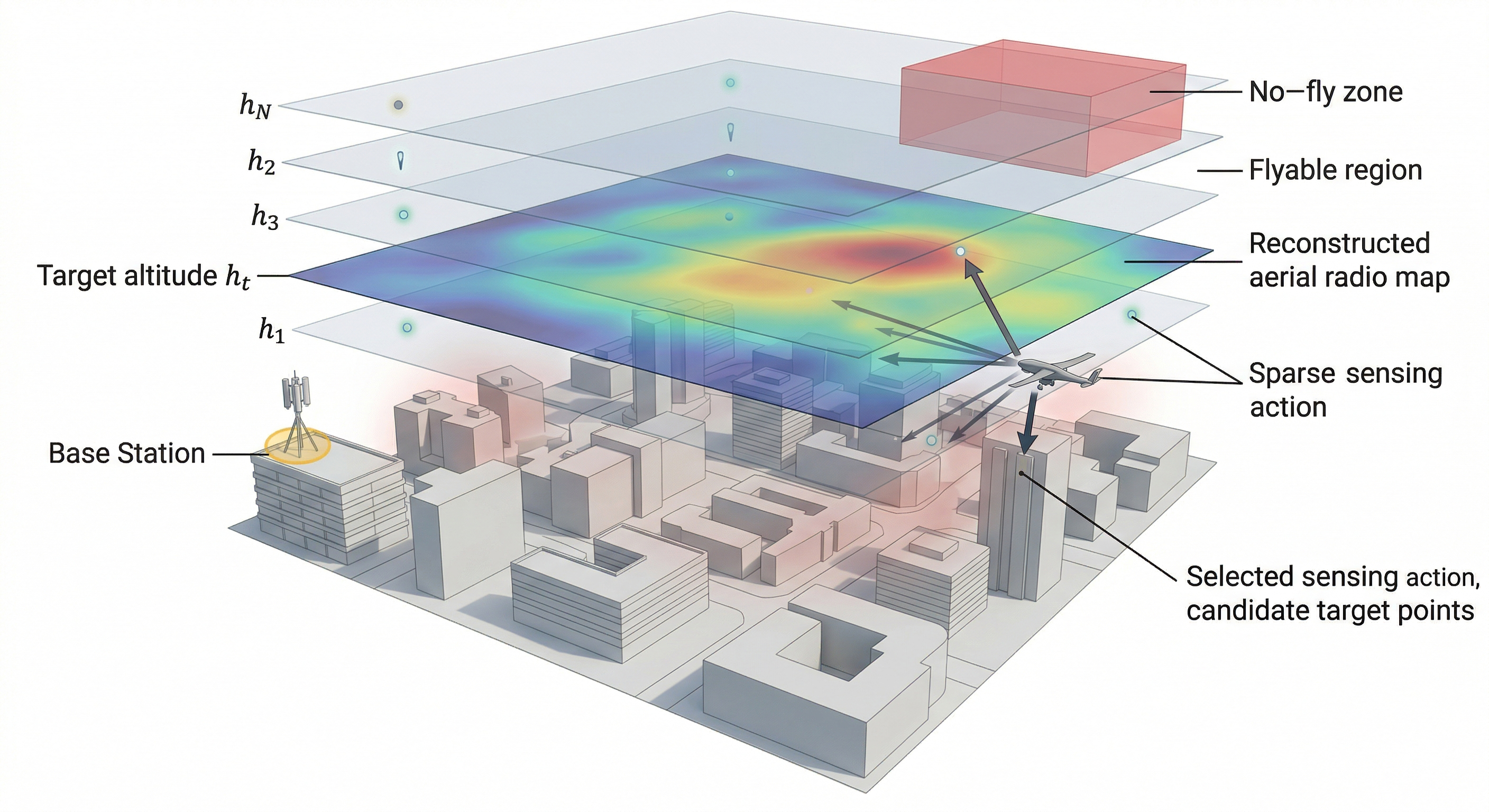}
 \caption{CKM prediction scenario from sparse measurements in a complex 3D urban environment.}
 \label{fig:architecture}
\end{figure}

We consider a low-altitude UAV communication scenario in a geometry-rich urban environment, as illustrated in Fig.~\ref{fig:architecture}, where a base station provides wireless coverage over a constrained aerial workspace. This setting is aligned with recent interest in aerial CKM construction, autonomous UAV surveying, and urban 3D propagation modeling \cite{Zeng2021SNARMUAV,Shrestha2023SpectrumSurveying,Chen2025Urban3DRadioMapSparse}. Different from conventional CKM modeling on a single ground plane, the objective here is to characterize and reconstruct a layered aerial CKM over multiple altitude levels. The key challenge is that the propagation field changes significantly with altitude because line-of-sight condition, blockage transition, and geometry-dependent multipath structure all evolve as the UAV moves vertically. Consequently, the channel-gain CKM slice at one altitude cannot be directly reused for another altitude, while exhaustive sensing over all altitude layers is infeasible under flight, safety, and motion constraints.

\subsection{Urban Geometry and Layered Aerial CKM Space}
\label{subsec_scene_model}

Let $\Omega \subset \mathbb{R}^{2}$ denote the horizontal service region, which is discretized into a regular grid with $N_x \times N_y$ cells. The urban environment contains buildings with different footprints and heights, and the base station is fixed at a three-dimensional coordinate. To model the low-altitude aerial workspace, we consider a finite altitude set $\mathcal{H}=\{h_1,h_2,\ldots,h_{N_h}\}$, where each altitude corresponds to one aerial CKM slice. The scene geometry and altitude partition are written as
\begin{align}
\Omega
&=
\left[-L_x,L_x\right]
\times
\left[-L_y,L_y\right]
\subset
\mathbb{R}^{2},
\\
\mathcal{G}
&=
\left\{
\mathbf{p}_{i,j}
=
\left(x_i,y_j\right)
\mid
i=1,\ldots,N_x,\ 
j=1,\ldots,N_y
\right\},
\\
h_{\mathrm{env}}\left(\mathbf{p}\right)
&=
\max_{1\le n\le N_b}
\left\{
H_n\,
\mathbf{1}
\left(
\mathbf{p}\in\mathcal{B}_n
\right)
\right\},
\\
m_{\mathrm{obs}}\left(\mathbf{p}\right)
&=
\mathbf{1}
\left(
h_{\mathrm{env}}\left(\mathbf{p}\right)>0
\right),
\\
\mathbf{s}_{\mathrm{B}}
&=
\left(
x_{\mathrm{B}},y_{\mathrm{B}},z_{\mathrm{B}}
\right),
\\
\mathbf{s}\left(\mathbf{p},h\right)
&=
\left(
x,y,h
\right),
\\
\mathcal{H}
&=
\mathcal{H}_{\mathrm{tr}}
\cup
\mathcal{H}_{\mathrm{in}}
\cup
\mathcal{H}_{\mathrm{ex}}.
\end{align}

Here $\mathcal{B}_n$ and $H_n$ denote the footprint and height of the $n$th building, $h_{\mathrm{env}}(\mathbf{p})$ is the scene height at horizontal position $\mathbf{p}$, and $m_{\mathrm{obs}}(\mathbf{p})$ is the corresponding obstacle mask. The subsets $\mathcal{H}_{\mathrm{tr}}$, $\mathcal{H}_{\mathrm{in}}$, and $\mathcal{H}_{\mathrm{ex}}$ denote the training, interpolation, and extrapolation altitude sets.

For each altitude layer $h\in\mathcal{H}$, we define a dense aerial CKM whose value at grid point $\mathbf{p}_{i,j}$ is the channel gain between the base station and a UAV located at $\mathbf{s}(\mathbf{p}_{i,j},h)$. The layered CKM representation and several cross-height descriptors are
\begin{align}
\mathbf{G}^{\star}_h
&\in
\mathbb{R}^{N_y\times N_x},
\\
\left[
\mathbf{G}^{\star}_h
\right]_{j,i}
&=
g^{\star}
\left(
\mathbf{p}_{i,j},h
\right),
\\
\mathcal{M}^{\star}
&=
\left\{
\mathbf{G}^{\star}_h
\right\}_{h\in\mathcal{H}},
\\
\mathbf{a}_{i,j}
&=
\left[
g^{\star}\left(\mathbf{p}_{i,j},h_1\right),
g^{\star}\left(\mathbf{p}_{i,j},h_2\right),
\ldots,
g^{\star}\left(\mathbf{p}_{i,j},h_{N_h}\right)
\right]^{\mathsf{T}},
\\
\nu_{i,j}^{\star}
&=
\arg\max_{1\le n\le N_h}
g^{\star}
\left(
\mathbf{p}_{i,j},h_n
\right),
\\
h_{i,j}^{\star}
&=
h_{\nu_{i,j}^{\star}},
\\
\Delta g_{i,j}
&=
\max_{1\le n\le N_h}
g^{\star}
\left(
\mathbf{p}_{i,j},h_n
\right)
-
\min_{1\le n\le N_h}
g^{\star}
\left(
\mathbf{p}_{i,j},h_n
\right).
\end{align}

Here, the superscript $\star$ indicates the ground-truth CKM or gain value generated by the propagation simulator. Specifically, $\mathbf{G}^{\star}_h$ denotes the dense aerial CKM at altitude $h$, whose $(j,i)$th element is the channel gain $g^{\star}(\mathbf{p}_{i,j},h)$ at horizontal grid point $\mathbf{p}_{i,j}$ and altitude $h$. The set $\mathcal{M}^{\star}$ collects the dense CKMs over all altitude layers in $\mathcal{H}$. For each horizontal grid point $\mathbf{p}_{i,j}$, $\mathbf{a}_{i,j}$ represents the vertical gain profile across all considered altitudes. The index $\nu_{i,j}^{\star}$ indicates the altitude layer with the maximum channel gain at $\mathbf{p}_{i,j}$, and $h_{i,j}^{\star}$ is the corresponding best-gain altitude. Finally, $\Delta g_{i,j}$ denotes the vertical gain span, namely the difference between the maximum and minimum gains across altitude layers at the same horizontal location. These quantities show that aerial channel quality is inherently altitude-sensitive, which motivates height-conditioned cross-height prediction rather than direct reuse of a single-height CKM.

\begin{figure*}[t!]
 \centering
 \includegraphics[width=0.85\linewidth]{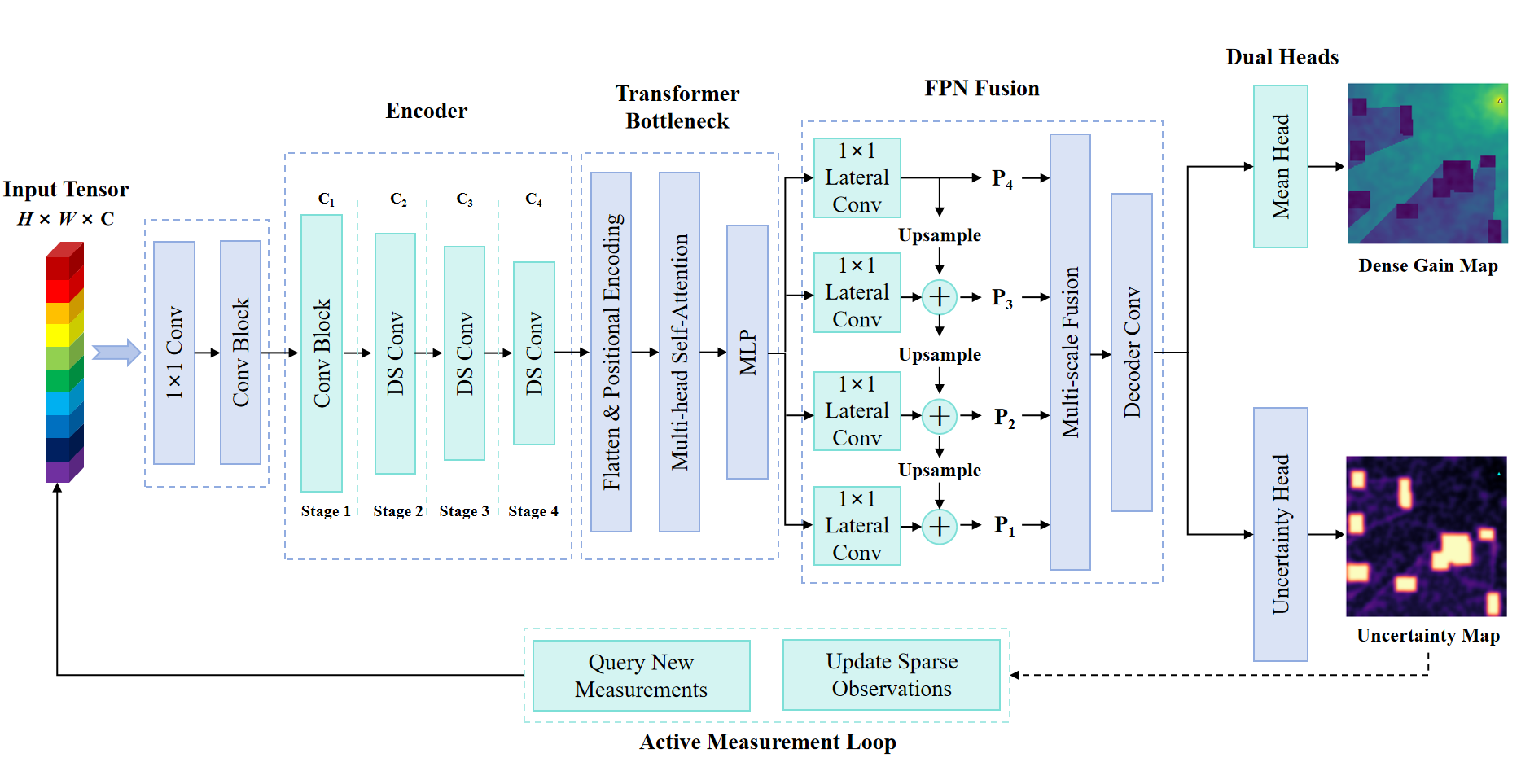}
 \caption{CKM reconstruction scenario from sparse measurements in a complex 3D urban environment.}
 \label{fig:model}
\end{figure*}

\subsection{Safe Flight Region and Feasible Aerial Sensing Space}
\label{subsec_flight_constraints}

In practical low-altitude UAV communication, measurements can only be collected from safe and reachable aerial states. The UAV must satisfy altitude limits, remain above the urban surface with sufficient safety clearance, and avoid no-fly zones. Moreover, even among feasible states, positions close to building boundaries or restricted areas may incur higher operational risk. Such execution-aware constraints are also consistent with prior UAV surveying and CKM-guided path planning studies \cite{Zeng2021SNARMUAV,Shrestha2023SpectrumSurveying,Mu2021IRSRobotPathPlanningRadioMap}. Let $h_{\min}$ and $h_{\max}$ denote the allowable altitude limits, let $\delta_{\mathrm{clr}}$ denote the required clearance margin, and let $\mathcal{Z}$ denote the set of no-fly zones. The feasibility and risk descriptors are
\begin{align}
c\left(\mathbf{p},h\right)
&=
h-h_{\mathrm{env}}\left(\mathbf{p}\right),
\\
m_{\mathrm{nfz}}\left(\mathbf{p},h\right)
&=
\mathbf{1}
\left(
\mathbf{p}\in\mathcal{Z}_h
\right),
\\
\begin{split}
m_{\mathrm{fly}}\left(\mathbf{p},h\right)
&=
\mathbf{1}\left(h_{\min}\le h\le h_{\max}\right)
\mathbf{1}\left(c\left(\mathbf{p},h\right)\ge\delta_{\mathrm{clr}}\right)\\
&\quad \times \left[1-m_{\mathrm{nfz}}\left(\mathbf{p},h\right)\right],
\end{split}
\\
d_{\mathrm{bd}}\left(\mathbf{p}\right)
&=
\min_{\mathbf{q}\in\partial\mathcal{B}}
\left\|
\mathbf{p}-\mathbf{q}
\right\|_2,
\\
r\left(\mathbf{p},h\right)
&=
\omega_1
\exp
\left(
-\frac{d_{\mathrm{bd}}\left(\mathbf{p}\right)}{\sigma_r}
\right)
+
\omega_2
m_{\mathrm{nfz}}\left(\mathbf{p},h\right),
\\
\omega_1+\omega_2
&=
1.
\end{align}

Here $c(\mathbf{p},h)$ is the vertical clearance above the urban surface, $m_{\mathrm{nfz}}(\mathbf{p},h)$ is the no-fly indicator, $m_{\mathrm{fly}}(\mathbf{p},h)$ is the flyable mask, $d_{\mathrm{bd}}(\mathbf{p})$ is the distance to the nearest building boundary, and $r(\mathbf{p},h)$ is the composite spatial risk field. Therefore, both observation collection and prediction evaluation are restricted to feasible aerial states.

\subsection{Cross-Height Sparse Observation and Prediction Task}
\label{subsec_sparse_observation_prediction}

Because exhaustive sensing over all altitude layers is prohibitively costly, the UAV only acquires sparse measurements from a limited set of observed heights and flyable positions. Let $\mathcal{H}_{\mathrm{obs}}\subseteq\mathcal{H}$ denote the observed altitude subset. At each observed altitude, a binary sampling mask determines which flyable grid points are sensed, and the resulting sparse CKM observation is
\begin{align}
\mathcal{H}_{\mathrm{obs}}
&=
\left\{
\tilde{h}_1,\tilde{h}_2,\ldots,\tilde{h}_K
\right\}
\subseteq
\mathcal{H},
\\
K
&\ll
N_h,
\\
\mathbf{M}_{\tilde{h}_k}
&\in
\left\{0,1\right\}^{N_y\times N_x},
\\
\left[
\mathbf{M}_{\tilde{h}_k}
\right]_{j,i}
&=
1
\Rightarrow
m_{\mathrm{fly}}
\left(
\mathbf{p}_{i,j},\tilde{h}_k
\right)
=
1,
\label{eq:obs_mask_flyable}
\\
\mathbf{Y}_{\tilde{h}_k}
&=
\mathbf{M}_{\tilde{h}_k}
\odot
\mathbf{G}^{\star}_{\tilde{h}_k}
+
\mathbf{M}_{\tilde{h}_k}
\odot
\mathbf{N}_{\tilde{h}_k},
\\
\mathcal{O}
&=
\left\{
\left(
\tilde{h}_k,
\mathbf{Y}_{\tilde{h}_k},
\mathbf{M}_{\tilde{h}_k}
\right)
\right\}_{k=1}^{K}.
\end{align}

Here, $\mathcal{H}_{\mathrm{obs}}$ is the subset of altitude layers where UAV measurements are available, and $\tilde{h}_k$ denotes the $k$th observed altitude. The number of observed altitude layers is $K$, which is much smaller than the total number of altitude layers $N_h$. The matrix $\mathbf{M}_{\tilde{h}_k}$ is a binary sampling mask at altitude $\tilde{h}_k$, where $\left[\mathbf{M}_{\tilde{h}_k}\right]_{j,i}=1$ means that the grid point $\mathbf{p}_{i,j}$ is measured at altitude $\tilde{h}_k$. The implication in \eqref{eq:obs_mask_flyable} enforces that measurements can only be collected from flyable aerial positions. The matrix $\mathbf{Y}_{\tilde{h}_k}$ denotes the sparse noisy CKM observation at altitude $\tilde{h}_k$, $\mathbf{N}_{\tilde{h}_k}$ denotes the corresponding measurement noise field, and $\odot$ denotes the element-wise product. The observation set $\mathcal{O}$ therefore contains all available measurement triplets, each consisting of an observed altitude, its sparse measurement CKM slice, and the associated sampling mask.

Given a target altitude $h_t\in\mathcal{H}$, the practical regime of interest is the cross-height case
\begin{align}
h_t \notin \mathcal{H}_{\mathrm{obs}},
\end{align}
where the dense CKM at the target height must be inferred from scene priors and sparse measurements collected at other heights. Let $\mathcal{P}(h_t)$ denote the geometry and target-height priors available at $h_t$, namely
\begin{align}
\mathcal{P}(h_t)
=
\left\{
\mathbf{H},
\mathbf{M}_{\mathrm{obs}},
\mathbf{R}_x,
\mathbf{R}_y,
\mathbf{D},
\mathbf{C}_{h_t},
\mathbf{F}_{h_t},
\mathbf{Z}_{h_t},
\mathbf{R}_{h_t}
\right\},
\end{align}
and let the valid target region be
\begin{align}
\mathbf{M}_{\mathrm{val}}(h_t)
=
\mathbf{F}_{h_t}
\odot
\left(
1-\mathbf{Z}_{h_t}
\right).
\label{eq:valid_mask_sys}
\end{align}
The learning task is then to estimate the dense target-height channel-gain CKM slice from $\mathcal{P}(h_t)$ and $\mathcal{O}$, a setting that has not been explicitly considered in previous UAV-assisted communication studies.

\subsection{Problem Formulation}
\label{subsec_problem_statement}

Let $\mathcal{T}_{\theta}(\cdot)$ denote a parametric predictor. Height-conditioned cross-height CKM prediction is formulated as
\begin{align}
\widehat{\mathbf{G}}_{h_t}
&=
\mathcal{T}_{\theta}
\left(
\mathcal{P}(h_t),
\mathcal{O},
h_t
\right),
\\
\min_{\theta}
\quad
&
\mathbb{E}
\left[
\mathcal{D}
\left(
\widehat{\mathbf{G}}_{h_t},
\mathbf{G}^{\star}_{h_t}
\right)
\right],
\end{align}
where $\mathcal{D}(\cdot,\cdot)$ denotes the reconstruction loss evaluated on the valid aerial region. In this paper, uncertainty estimation and cost-aware measurement selection are treated as downstream deployment modules built on top of this primary prediction task, rather than as co-primary formulations.

\section{Methodology}
\label{sec_method}

\subsection{Overview of the Proposed Framework}
\label{subsec_method_overview}

The proposed framework is centered on height-conditioned dense CKM prediction as shown in Fig.~\ref{fig:model}. Given scene priors, sparse multi-height observations, and a queried target height, the model reconstructs the dense channel-gain CKM slice at that target layer. On top of this primary predictor, an auxiliary uncertainty head is learned to characterize prediction confidence and to support downstream online sensing.

Accordingly, the methodology contains four connected components. First, geometry priors, height-indexed sparse observations, and target-height descriptors are organized into a unified input tensor. Second, an FPN-Transformer backbone extracts hierarchical spatial features and reconstructs the dense target-height channel-gain CKM slice. Third, an uncertainty head predicts a confidence map over the valid aerial region. Fourth, the predicted uncertainty is converted into a cost-aware sensing utility under flight cost and risk constraints. This organization keeps prediction as the main methodological focus, while treating active sensing as a downstream extension of the learned predictor.

\subsection{Height-Conditioned Input Construction}
\label{subsec_input_construction}

The input representation is designed to remain structurally consistent across different observation patterns while explicitly identifying the queried target height. Let the reserved training height set be denoted by $\mathcal{H}_{\mathrm{tr}}=\{h_1,h_2,\ldots,h_K\}$, where each $h_k$ corresponds to one fixed observation slot in the network input. For a target height $h_t$, the complete tensor $\mathbf{X}(h_t)$ is formed by concatenating a static scene block, a sequence of height-indexed observation blocks, and a target-specific descriptor block given by:

\begin{align}
\mathbf{X}(h_t) &=
\left[
\mathbf{X}_{\mathrm{sta}},
\mathbf{X}_{\mathrm{obs}}^{(1)},
\mathbf{X}_{\mathrm{obs}}^{(2)},
\ldots,
\mathbf{X}_{\mathrm{obs}}^{(K)},
\mathbf{X}_{\mathrm{tar}}(h_t)
\right], \label{eq:full_input_tensor_new} \\
\mathbf{X}_{\mathrm{obs}}^{(k)} &=
\begin{cases}
\left[
\mathbf{Y}_{h_k},
\mathbf{M}_{h_k},
\mathbf{A}_{h_k}
\right], & h_k \in \mathcal{H}_{\mathrm{obs}},\\
\mathbf{0}, & h_k \notin \mathcal{H}_{\mathrm{obs}}.
\end{cases}
\label{eq:obs_slot_new}
\end{align}

Here $\mathbf{X}_{\mathrm{sta}}$ contains scene-level descriptors shared across height layers, such as scene height distribution, obstacle layout, coordinate offsets, and distance fields. The target block $\mathbf{X}_{\mathrm{tar}}(h_t)$ contains descriptors of the queried layer, including target height condition, flyable mask, clearance cue, line-of-sight cue, spatial risk, and no-fly information. To make the queried height comparable across layers, a normalized target-height descriptor is introduced as
\begin{align}
\eta(h_t) &=
\frac{h_t-h_{\min}^{\mathcal{H}}}
{h_{\max}^{\mathcal{H}}-h_{\min}^{\mathcal{H}}}, \\
h_{\min}^{\mathcal{H}} &= \min_{h\in\mathcal{H}} h, \\
h_{\max}^{\mathcal{H}} &= \max_{h\in\mathcal{H}} h.
\label{eq:height_norm_new}
\end{align}

The valid target region is given by $\mathbf{M}_{\mathrm{val}}(h_t)$ in \eqref{eq:valid_mask_sys}. In addition, to stabilize optimization and suppress the influence of extreme gain values, the ground-truth channel-gain CKM slice is clipped to a fixed dynamic range and normalized before training:
\begin{align}
\bar{\mathbf{G}}_{h_t}^{\star}
&=
\frac{
\operatorname{clip}\!\left(\mathbf{G}_{h_t}^{\star},G_{\min},G_{\max}\right)-G_{\min}
}{
G_{\max}-G_{\min}
}.
\label{eq:gain_norm_new}
\end{align}

\subsection{FPN-Transformer-Based Cross-Height Predictor}
\label{subsec_fpn_transformer_predictor}

The core predictor is built upon an FPN-Transformer architecture, which combines the local modeling ability of convolutional encoding with the global dependency modeling capability of self-attention. This design is also broadly consistent with recent CKM estimators that leverage convolutional recovery, graph or deformable attention, generative refinement, and efficient backbone design \cite{Levie2021RadioUNet,Li2024RadioGAT,Liu2026DATUnetRadioMap,Wang2025RadioDiff}. This choice matches the target problem well. Local blockage boundaries, street corridors, and structural transitions require fine spatial detail preservation, whereas cross-height inference also depends on broader scene context because the gain distribution at the queried layer is determined by the interaction between target height, surrounding geometry, and sparse observations collected at other heights.

Given the input tensor $\mathbf{X}(h_t)$, the network first performs four-stage convolutional encoding to generate hierarchical features at progressively coarser resolutions:
\begin{align}
\mathbf{C}_1 &= \phi_1\!\left(\mathbf{X}(h_t)\right), \\
\mathbf{C}_2 &= \phi_2\!\left(\mathbf{C}_1\right), \\
\mathbf{C}_3 &= \phi_3\!\left(\mathbf{C}_2\right), \\
\mathbf{C}_4 &= \phi_4\!\left(\mathbf{C}_3\right).
\label{eq:encoder_group_new}
\end{align}
Here $\phi_1(\cdot)$ to $\phi_4(\cdot)$ denote convolutional encoding operators, and the later stages use stride-based downsampling to enlarge the receptive field. The deepest feature $\mathbf{C}_4$ is reshaped into a token sequence and refined through a transformer block:
\begin{align}
\mathbf{Z} &= \operatorname{Reshape}\!\left(\mathbf{C}_4\right)\in\mathbb{R}^{B\times HW\times C}, \\
\tilde{\mathbf{Z}} &= \mathcal{T}\!\left(\mathbf{Z}\right), \\
\tilde{\mathbf{C}}_4 &= \operatorname{Reshape}^{-1}\!\left(\tilde{\mathbf{Z}}\right).
\label{eq:transformer_group_new}
\end{align}
Here $\mathcal{T}(\cdot)$ denotes the transformer refinement module composed of normalization, multi-head self-attention, feed-forward transformation, and residual connection.

After global refinement, a top-down feature pyramid fuses deep semantic features with shallow spatial cues:
\begin{align}
\mathbf{P}_4 &= \psi_4\!\left(\tilde{\mathbf{C}}_4\right), \\
\mathbf{P}_3 &= \psi_3\!\left(\mathbf{C}_3\right)+\operatorname{Up}\!\left(\mathbf{P}_4\right), \\
\mathbf{P}_2 &= \psi_2\!\left(\mathbf{C}_2\right)+\operatorname{Up}\!\left(\mathbf{P}_3\right), \\
\mathbf{P}_1 &= \psi_1\!\left(\mathbf{C}_1\right)+\operatorname{Up}\!\left(\mathbf{P}_2\right), \\
\mathbf{F}_{\mathrm{fuse}} &=
f_{\mathrm{fuse}}
\left(
\left[
\operatorname{Up}\!\left(\mathbf{P}_1\right),
\operatorname{Up}\!\left(\mathbf{P}_2\right),
\operatorname{Up}\!\left(\mathbf{P}_3\right),
\operatorname{Up}\!\left(\mathbf{P}_4\right)
\right]
\right).
\label{eq:fpn_group_new}
\end{align}
Here $\psi_1(\cdot)$ to $\psi_4(\cdot)$ denote lateral projection layers and $\operatorname{Up}(\cdot)$ denotes spatial upsampling. The fused representation $\mathbf{F}_{\mathrm{fuse}}$ preserves both global propagation context and local geometric detail, which is essential for cross-height dense reconstruction from sparse observations.

\subsection{Gain Estimation and Uncertainty Learning}
\label{subsec_gain_uncertainty_learning}

Based on the shared fused feature $\mathbf{F}_{\mathrm{fuse}}$, the network uses two task-specific output heads to estimate the target channel-gain CKM slice and its associated uncertainty. The outputs are
\begin{align}
\hat{\bar{\mathbf{G}}}_{h_t} &= \sigma\!\left(f_g\!\left(\mathbf{F}_{\mathrm{fuse}}\right)\right), \\
\hat{\mathbf{U}}_{h_t} &= \rho\!\left(f_u\!\left(\mathbf{F}_{\mathrm{fuse}}\right)\right), \\
\hat{\mathbf{G}}_{h_t} &= \hat{\bar{\mathbf{G}}}_{h_t}\left(G_{\max}-G_{\min}\right)+G_{\min}.
\label{eq:output_group_new}
\end{align}
Here $\sigma(\cdot)$ denotes the sigmoid mapping used for normalized gain prediction, and $\rho(\cdot)$ denotes a nonnegative activation for uncertainty estimation.

Since direct uncertainty supervision is generally unavailable in CKM prediction, we construct a teacher uncertainty map from target-height descriptors and scene-level difficulty cues. Intuitively, regions associated with higher risk, weaker line-of-sight conditions, and larger cross-height gain variations are more difficult to predict reliably from sparse observations. Based on this intuition, the teacher uncertainty map is defined as
\begin{align}
\mathbf{U}_{h_t}^{\star}
&=
\mathbf{M}_{\mathrm{val}}(h_t)\odot
\left[
\alpha\,\mathbf{P}_{h_t}^{\mathrm{risk}}
+
\beta\,\left(1-\mathbf{L}_{h_t}\right)
+
\gamma\,\Delta\mathbf{G}
\right],
\label{eq:teacher_uncertainty_new}
\end{align}
where $\alpha$, $\beta$, and $\gamma$ denote the weighting coefficients for the risk map, the line-of-sight deficiency term, and the cross-height gain span descriptor, respectively. Here $\mathbf{P}_{h_t}^{\mathrm{risk}}$ denotes the target-height risk map, $\mathbf{L}_{h_t}$ denotes the target-height line-of-sight cue, and $\Delta\mathbf{G}$ denotes the cross-height gain span descriptor. Before fusion, the three cue maps are normalized to a comparable range, and the coefficients are constrained to satisfy $\alpha+\beta+\gamma=1$ for interpretability. In this work, we set $\alpha=0.45$, $\beta=0.30$, and $\gamma=0.25$ according to their relative importance in low-altitude UAV sensing. The risk term is assigned the largest weight because regions close to obstacles or no-fly boundaries are both operationally critical and more difficult to sample safely. The line-of-sight deficiency term is assigned the second largest weight since blockage transitions are a major source of prediction uncertainty in urban propagation. The cross-height gain span term is used as an auxiliary cue to capture vertical variation, but with a smaller weight to avoid overemphasizing altitude-wise gain fluctuation at the expense of safety and visibility information. These coefficients are fixed for all experiments and are not tuned separately for different test scenes.

\subsection{Training Objective and Rapid Deployment Strategy}
\label{subsec_training_and_deployment}

The overall training objective is defined on the valid aerial region and balances pointwise fidelity, local structural consistency, and uncertainty supervision. Let $\hat{\bar{\mathbf{G}}}_{h_t}$ and $\bar{\mathbf{G}}_{h_t}^{\star}$ denote the predicted and ground-truth normalized channel-gain CKM slices. The loss terms are
\begin{align}
\mathcal{L}_{\ell_1}
&=
\frac{
\left\|
\mathbf{M}_{\mathrm{val}}\odot
\left(
\hat{\bar{\mathbf{G}}}_{h_t}-\bar{\mathbf{G}}_{h_t}^{\star}
\right)
\right\|_1
}{
\left\|\mathbf{M}_{\mathrm{val}}\right\|_1
}, \\
\mathcal{L}_{\mathrm{char}}
&=
\frac{
\sum_{i,j}
M_{\mathrm{val},i,j}
\sqrt{
\left(
\hat{\bar{G}}_{i,j}-\bar{G}_{i,j}^{\star}
\right)^2+\epsilon^2
}
}{
\sum_{i,j} M_{\mathrm{val},i,j}
}, \\
\mathcal{L}_{\mathrm{grad}}
&=
\frac{1}{2}
\left(
\mathcal{L}_{x}^{\mathrm{grad}}+\mathcal{L}_{y}^{\mathrm{grad}}
\right), \\
\mathcal{L}_{u}
&=
\frac{
\left\|
\mathbf{M}_{\mathrm{val}}\odot
\left(
\hat{\mathbf{U}}_{h_t}-\mathbf{U}_{h_t}^{\star}
\right)
\right\|_1
}{
\left\|\mathbf{M}_{\mathrm{val}}\right\|_1
}.
\label{eq:loss_group_new}
\end{align}
Here $\mathcal{L}_{x}^{\mathrm{grad}}$ and $\mathcal{L}_{y}^{\mathrm{grad}}$ denote the gradient discrepancies along the horizontal and vertical directions, respectively. The final objective is
\begin{align}
\mathcal{L}
=
\lambda_{\ell_1}\mathcal{L}_{\ell_1}
+
\lambda_{\mathrm{char}}\mathcal{L}_{\mathrm{char}}
+
\lambda_{\mathrm{grad}}\mathcal{L}_{\mathrm{grad}}
+
\lambda_{\mathrm{u}}\mathcal{L}_{u}.
\label{eq:total_loss_new}
\end{align}
Here $\lambda_{\ell_1}$, $\lambda_{\mathrm{char}}$, $\lambda_{\mathrm{grad}}$, and $\lambda_{\mathrm{u}}$ are the weights of the masked $\ell_1$ loss, Charbonnier loss, gradient consistency loss, and uncertainty supervision loss, respectively. In all experiments, we set $\lambda_{\ell_1}=1.0$, $\lambda_{\mathrm{char}}=0.6$, $\lambda_{\mathrm{grad}}=0.15$, and $\lambda_{\mathrm{u}}=0.10$.

\begin{algorithm}[!t]
\caption{Height-Conditioned CKM Prediction}
\label{alg:method_uav}
\begin{algorithmic}[1]
\REQUIRE Geometry priors $\mathcal{P}$, initial sparse observations $\mathcal{O}^{(0)}$, target height $h_t$, initial UAV state $\mathbf{s}_0$, query budget $K_q$, active iteration number $T$, trained network $\mathcal{F}_{\theta}$
\ENSURE Final reconstructed target CKM slice $\hat{\mathbf{G}}_{h_t}^{(T_f)}$, final uncertainty map $\hat{\mathbf{U}}_{h_t}^{(T_f)}$, sensing cost $\Gamma^{(T_f)}$

\STATE $\Gamma^{(0)} \leftarrow [\mathbf{s}_0]$
\STATE $\mathbf{X}^{(0)}(h_t) \leftarrow \Pi\!\left(\mathcal{P}, \mathcal{O}^{(0)}, h_t\right)$
\STATE $\mathbf{M}_{\mathrm{val}} \leftarrow \mathbf{F}_{h_t} \odot \left(1 - \mathbf{Z}_{h_t}\right)$

\FOR{$t = 0$ to $T - 1$}
 \STATE $\left(\hat{\bar{\mathbf{G}}}_{h_t}^{(t)}, \hat{\mathbf{U}}_{h_t}^{(t)}\right) \leftarrow \mathcal{F}_{\theta}\!\left(\mathbf{X}^{(t)}(h_t)\right)$
 \STATE $\hat{\mathbf{G}}_{h_t}^{(t)} \leftarrow \hat{\bar{\mathbf{G}}}_{h_t}^{(t)} \left(G_{\max} - G_{\min}\right) + G_{\min}$
 \STATE $\Omega^{(t)} \leftarrow \left\{ a \in \mathcal{A}\!\left(\mathbf{s}_t\right) \mid a \notin \mathcal{O}^{(t)} \right\}$
 \IF{$\Omega^{(t)} = \varnothing$}
 \STATE \textbf{break}
 \ENDIF
 \STATE $\Psi^{(t)}(a) \leftarrow \left[\hat{\mathbf{U}}_{h_t}^{(t)}\right]_{a} - \beta C_{\mathrm{tot}}\!\left(a \mid \mathbf{s}_t\right), \quad \forall a \in \Omega^{(t)}$
 \STATE $\mathcal{Q}^{(t)} \leftarrow \operatorname{TopK}\!\left(\Psi^{(t)}, K_q\right)$
 \STATE $\mathcal{O}^{(t+1)} \leftarrow \mathcal{O}^{(t)} \cup \left\{ \left(a, G(a)\right) \mid a \in \mathcal{Q}^{(t)} \right\}$
 \STATE $\mathbf{s}_{t+1} \leftarrow \operatorname{Move}\!\left(\mathbf{s}_t, \mathcal{Q}^{(t)}\right)$
 \STATE $\Gamma^{(t+1)} \leftarrow \Gamma^{(t)} \oplus \mathbf{s}_{t+1}$
 \STATE $\mathbf{X}^{(t+1)}(h_t) \leftarrow \Pi\!\left(\mathcal{P}, \mathcal{O}^{(t+1)}, h_t\right)$
\ENDFOR

\STATE $T_f \leftarrow \min\!\left(t+1, T\right)$
\STATE \textbf{return} $\hat{\mathbf{G}}_{h_t}^{(T_f)}, \hat{\mathbf{U}}_{h_t}^{(T_f)}, \Gamma^{(T_f)}$
\end{algorithmic}
\end{algorithm}

The deployment protocol follows emergency UAV sensing conditions, where the model may be applied immediately in a new scene or adapted from a very small local support set. In zero-shot deployment, the trained model is directly applied to an unseen scene:
\begin{align}
\left(
\hat{\mathbf{G}}_{h_t},
\hat{\mathbf{U}}_{h_t}
\right)
=
\mathcal{F}_{\theta}\!\left(\mathbf{X}(h_t)\right).
\label{eq:zero_shot_new}
\end{align}
In few-shot deployment, a small support set from the new scene is used for rapid adaptation before query prediction:
\begin{align}
\theta^{\prime}
=
\arg\min_{\vartheta}
\frac{1}{|\mathcal{S}|}
\sum_{(\mathbf{X},\mathbf{G}^{\star},\mathbf{U}^{\star})\in\mathcal{S}}
\mathcal{L}\!\left(\vartheta;\mathbf{X},\mathbf{G}^{\star},\mathbf{U}^{\star}\right).
\label{eq:few_shot_new}
\end{align}

\subsection{Uncertainty-Guided Cost-Aware Sensing}
\label{subsec_uncertainty_guided_sensing}

After the dense channel-gain CKM slice and uncertainty map are obtained, the predictor is further coupled with an online sensing policy. In low-altitude UAV sensing, the cost of acquiring a new measurement depends not only on planar motion, but also on altitude transition and spatial risk. Therefore, the next action should be selected by jointly considering uncertainty benefit and travel burden.

Let the current UAV state be $\mathbf{s}_c=\left(\mathbf{p}_c,h_c\right)$ and let $a=\left(\mathbf{p},h\right)$ denote a candidate sensing action. The motion cost model and feasible action set are
\begin{align}
C_{\mathrm{hor}}\left(a \mid \mathbf{s}_c\right)
&=
\left\|
\mathbf{p}-\mathbf{p}_c
\right\|_2,
\\
C_{\mathrm{ver}}\left(a \mid \mathbf{s}_c\right)
&=
\left|
h-h_c
\right|,
\\
C_{\mathrm{tot}}\left(a \mid \mathbf{s}_c\right)
&=
\lambda_1
C_{\mathrm{hor}}\left(a \mid \mathbf{s}_c\right)
+
\lambda_2
C_{\mathrm{ver}}\left(a \mid \mathbf{s}_c\right)
+
\lambda_3
r\left(\mathbf{p},h\right),
\\
\mathcal{A}\left(\mathbf{s}_c\right)
&=
\left\{
a=\left(\mathbf{p},h\right)
\mid
m_{\mathrm{fly}}\left(\mathbf{p},h\right)=1
\right\}.
\end{align}

Using the predicted uncertainty as a surrogate for information gain, the sensing utility is defined as
\begin{align}
\Psi(a)
&=
\left[\hat{\mathbf{U}}_{h}\right]_{j,i}
-
\beta C_{\mathrm{tot}}\!\left(a\mid\mathbf{s}_c\right), \\
a^{\star}
&=
\arg\max_{a\in\mathcal{A}(\mathbf{s}_c)}
\Psi(a), \\
\mathcal{Q}^{(t)}
&=
\operatorname{TopK}
\left(
\hat{\mathbf{U}}^{(t)}\odot\mathbf{M}_{\mathrm{val}}^{(t)}
-
\beta \mathbf{C}^{(t)}
\right),
\quad a=\left(\mathbf{p}_{i,j},h\right).
\label{eq:sensing_group_new}
\end{align}

Here the second equation selects the best single action, whereas the third equation gives a top-ranked candidate subset for batch acquisition. For comparison, three baseline sensing policies are considered on the same feasible action set:
\begin{align}
a^{\mathrm{rand}} &\sim \mathrm{Uniform}\big(\mathcal{A}(\mathbf{s}_c)\big), \\
a^{\mathrm{near}} &= \arg\min_{a \in \mathcal{A}(\mathbf{s}_c)} C_{\mathrm{tot}}(a \mid \mathbf{s}_c), \\
a^{\mathrm{unc}} &= \arg\max_{a \in \mathcal{A}(\mathbf{s}_c)} \left[\hat{\mathbf{U}}_{h}\right]_{j,i},
\quad a=\left(\mathbf{p}_{i,j},h\right).
\end{align}
These baselines correspond to random aerial sampling, nearest greedy travel, and uncertainty-only sensing, respectively. The proposed policy is given by \eqref{eq:sensing_group_new}. In this way, the same network output that improves CKM prediction is also used to guide practical measurement selection under motion and risk constraints.

\section{Experimental Setup}
\label{sec_experiments}

\subsection{Dataset Construction}
\label{subsec_dataset_construction}

The experiments are conducted on a layered aerial CKM benchmark generated specifically for height-conditioned cross-height prediction. Different from the current three-scene setting, the benchmark is expanded to a more diverse controlled urban scene collection in order to provide stronger evidence for unseen-scene generalization. In total, we construct seven scene instances, including four road-oriented layouts and three non-road building layouts. The road-oriented group contains a crossroad scene, a T-junction scene, a canyon scene, and an offset-crossroad scene. The non-road group contains dense, medium, and sparse building layouts. This design allows us to jointly vary road topology, building density, and spatial blockage complexity, while also making the cross-height evaluation less dependent on a very small scene pool. It therefore provides a more convincing testbed for studying sparse multi-altitude prediction under scene shift, and is also consistent with recent interest in benchmark-oriented CKM research \cite{Xu2024HowMuchDataCKM,Wang2025RadioDiff,Chen2025Urban3DRadioMapSparse}.

\begin{figure}[!t]
\centering
\subfloat[3D BS and UAV scene.]{
\includegraphics[height=2.4cm, keepaspectratio]{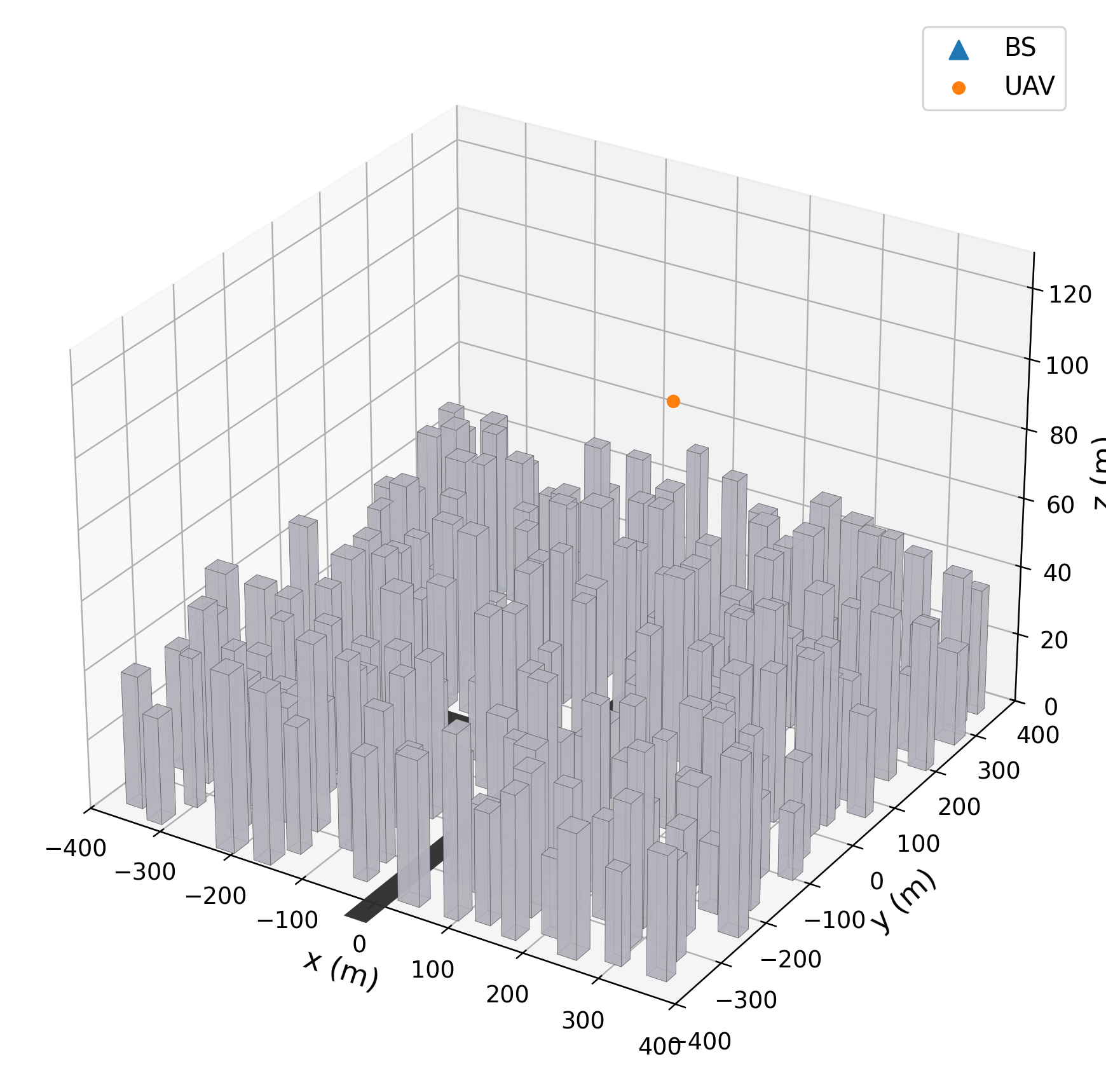}
\label{fig:dataset_scene_example_3d}}
\hfill
\subfloat[2D urban layout.]{
\includegraphics[height=2.4cm, keepaspectratio]{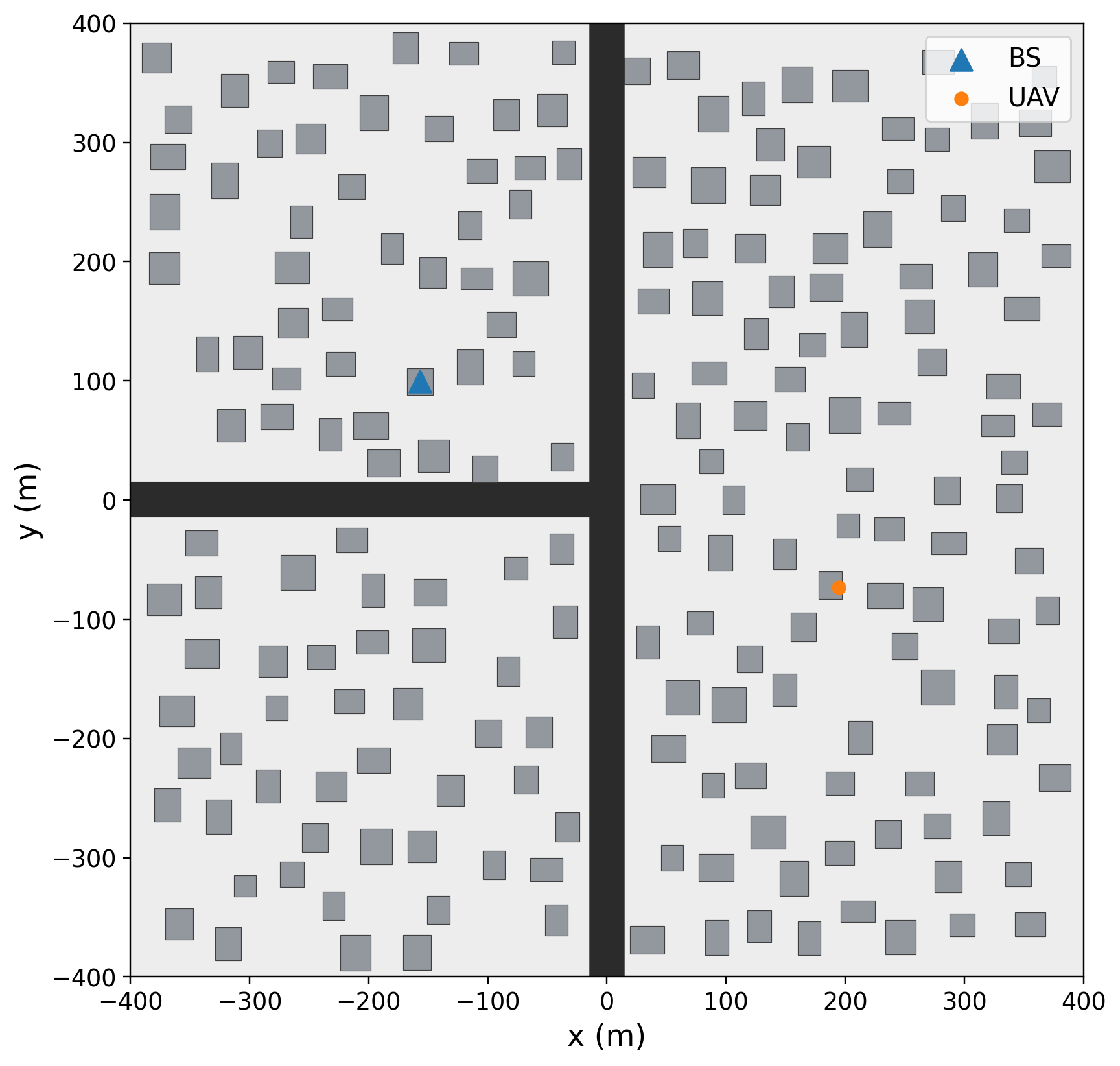}
\label{fig:dataset_scene_example_layout}}
\hfill
\subfloat[Channel-gain CKM slice at target UAV height.]{
\includegraphics[height=2.4cm, keepaspectratio]{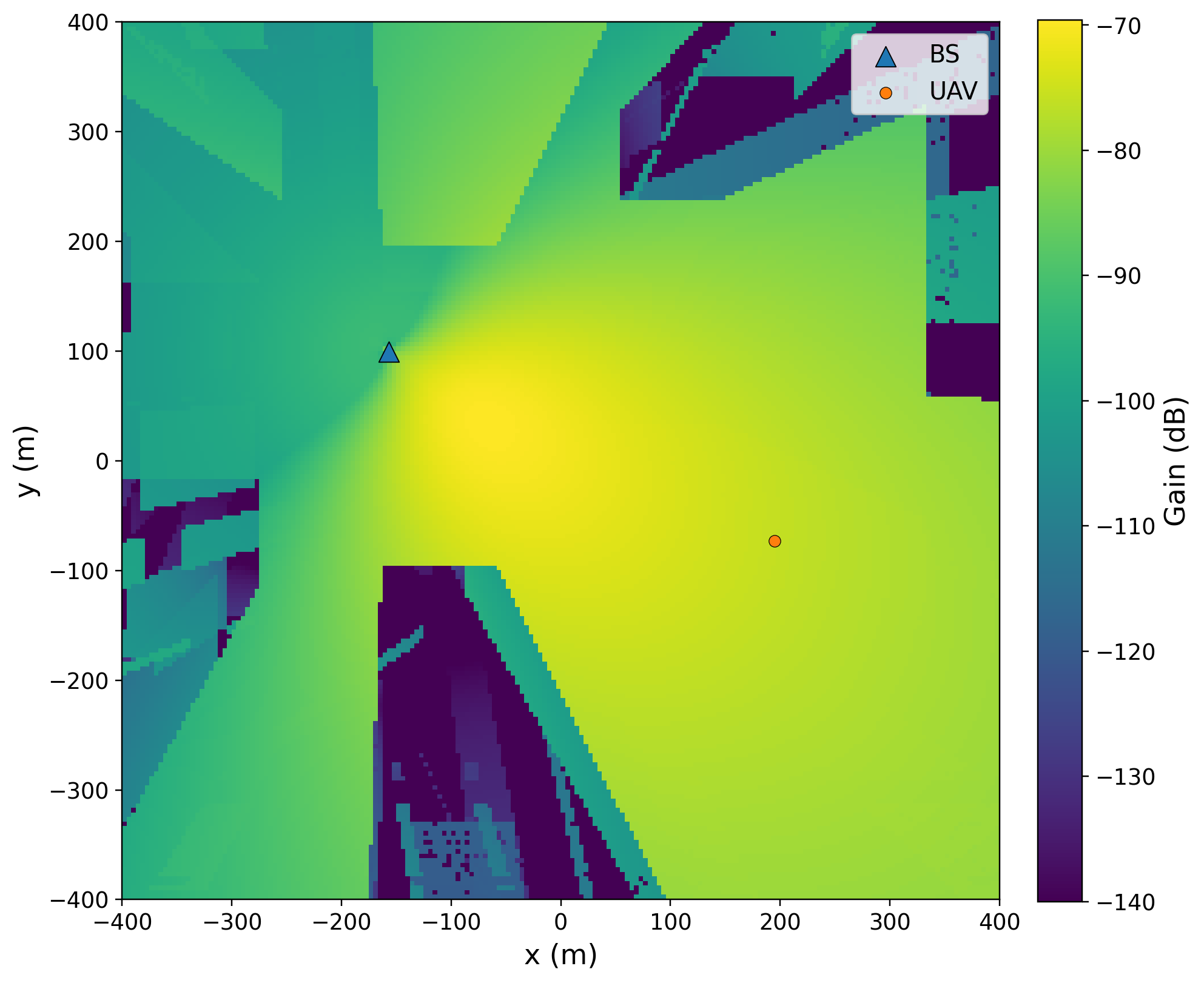}
\label{fig:dataset_scene_example_gain}}
\caption{Representative example of the proposed UAV CKM benchmark.}
\label{fig:dataset_scene_example}
\end{figure}

For each scene, buildings are randomly generated under geometric constraints on width, length, height, and inter-building clearance. More specifically, the horizontal service extent is set to $400\,\mathrm{m}$, the global map resolution is $192 \times 192$, and the cropped patch size is $96 \times 96$. The street width is fixed to $30\,\mathrm{m}$. Depending on the scene category, the number of buildings varies from $120$ to $300$, producing low-density, medium-density, and high-density propagation environments. The building width and length are sampled from $18\,\mathrm{m}$ to $30\,\mathrm{m}$, the building height is sampled from $18\,\mathrm{m}$ to $55\,\mathrm{m}$, and the minimum inter-building clearance is set to $8\,\mathrm{m}$. Each scene contains $4$ base-station deployments drawn from roadside and rooftop modes, which yields diverse transmitter positions and viewing geometries while keeping the aerial benchmark size manageable. For each valid transmitter, $48$ patches are extracted after enforcing patch-level filtering constraints on building coverage and valid gain support, so that each patch contains sufficient geometric and propagation diversity for cross-height reconstruction. Fig.~\ref{fig:dataset_scene_example} presents a representative example of the generated environment, including the three-dimensional urban scene, the corresponding two-dimensional layout, and the target-height channel-gain field.

\begin{table}[!t]
\centering
\caption{Main dataset configuration}
\label{tab:dataset_config_uav}
\setlength{\tabcolsep}{18pt}
\begin{tabular}{ll}
\toprule
\textbf{Item} & \textbf{Setting} \\
\midrule

Scene number & $7$ \\
Buildings per scene & $120$ to $300$ \\
BS number per scene & $4$ \\
BS modes & Roadside, Rooftop \\
Horizontal extent & $400\,\mathrm{m}$ \\
Street width & $30\,\mathrm{m}$ \\
Global resolution & $192 \times 192$ \\
Patch size & $96 \times 96$ \\
Patches per BS & $48$ \\
Carrier frequency & $3.5\,\mathrm{GHz}$ \\
Ray tracing depth & $3$ \\
Ray tracing samples & $1.2 \times 10^{5}$ \\
UAV heights & $20, 60, 100, 120\,\mathrm{m}$ \\
Training heights & $20, 60, 100\,\mathrm{m}$ \\
Extrapolation height & $120\,\mathrm{m}$ \\
Sampling ratios & $5\%, 10\%, 20\%$ \\
No-fly zone count & $1$ to $2$ \\
No-fly zone size & $40\,\mathrm{m}$ to $85\,\mathrm{m}$ \\
No-fly height limit & $40\,\mathrm{m}$ to $100\,\mathrm{m}$ \\
\bottomrule
\end{tabular}
\end{table}

Dense channel-gain CKM slices are generated by Sionna RT under a unified propagation configuration \cite{hoydis2023sionnart}. The carrier frequency is set to $3.5\,\mathrm{GHz}$, the ray-tracing depth is limited to $3$, and $1.2 \times 10^{5}$ tracing samples are used for dense field generation. For each scene and base-station configuration, the benchmark further provides geometry-aware side information channels, including height maps, obstacle masks, relative coordinate fields, transmitter distance maps, flyable-region masks, target-height descriptors, no-fly indicators, risk fields, and cross-height profile cues. The sparse CKM observations are then synthesized at multiple heights and organized into the height-conditioned input format described in Section~\ref{sec_method}-B.

The layered aerial CKM setting is built on four UAV heights, namely $20\,\mathrm{m}$, $60\,\mathrm{m}$, $100\,\mathrm{m}$, and $120\,\mathrm{m}$. Among them, $20\,\mathrm{m}$, $60\,\mathrm{m}$, and $100\,\mathrm{m}$ are used as training heights, whereas $120\,\mathrm{m}$ is reserved as the extrapolation height. The minimum safe altitude is set to $20\,\mathrm{m}$, the maximum safe altitude is set to $120\,\mathrm{m}$, and the clearance margin is set to $4\,\mathrm{m}$. To evaluate the effect of measurement support under aerial constraints, sparse observations are generated under three observation ratios, namely $5\%$, $10\%$, and $20\%$, and under three mask types, namely random flyable sampling, boundary-aware sampling, and travel-cost-aware sampling. The initialization CKM slice is generated by iterative mean filling.

To reflect realistic aerial deployment constraints, the benchmark also includes no-fly regions and risk fields. The number of no-fly zones is randomly selected from $1$ to $2$, the zone size ranges from $40\,\mathrm{m}$ to $85\,\mathrm{m}$, and the associated height limit ranges from $40\,\mathrm{m}$ to $100\,\mathrm{m}$. The risk field is controlled by a decay parameter $\sigma = 28\,\mathrm{m}$, with boundary weight $0.55$ and no-fly weight $0.45$. In addition, trajectory-related cost descriptors are generated using horizontal, vertical, and risk weights equal to $1.0$, $1.2$, and $0.8$, respectively. For each patch, the sequential metadata stores up to $4$ sensing steps and a candidate set with at most $256$ points, which supports subsequent analysis of cost-aware uncertainty-guided sensing.

The benchmark provides two complementary evaluation protocols. The first protocol is a scene-level generalization protocol designed to test genuine unseen-scene transfer. In this protocol, the dataset is first split by scene identity, where five scenes are used for model development and two scenes are held out for unseen-scene testing. Within the seen-scene subset, the sample units are further divided into training and validation subsets with ratios $0.80$ and $0.20$. Compared with a three-scene split, this protocol yields a substantially stronger and more stable evaluation of cross-scene generalization because the held-out test set is no longer dominated by a single scene type. The second protocol is a legacy patch-random split, where all sample units are randomly partitioned into training, validation, and test subsets with ratios $0.70$, $0.15$, and $0.15$. This protocol is retained to show the gap between conventional same-distribution evaluation and the more challenging unseen-scene setting.

To support rapid deployment analysis, the benchmark further defines a few-shot support-query protocol on unseen scenes. The support budget is selected from $\{0,1,5,10\}$, and the number of support heights is selected from $\{1,2\}$. For each unseen scene and each budget setting, the support samples and query samples are recorded explicitly. As a result, the same benchmark supports zero-shot evaluation, few-shot adaptation, cross-height extrapolation analysis, and cost-aware active sensing analysis within a unified experimental framework.

\subsection{Implementation Details}
\label{subsec_implementation_details}

\subsubsection{Hardware and Software Environment}

All computations were carried out on a workstation configured with an Intel Core i9 12900KF processor, an NVIDIA GeForce RTX 5060 Ti graphics card with $16\,\mathrm{GB}$ memory, and $64\,\mathrm{GB}$ system random access memory (RAM). The layered aerial CKM benchmark was generated in Python 3.11 by using Sionna RT together with TensorFlow. The learning-based reconstruction models were implemented in PyTorch 2.x. Graphics processing unit (GPU) acceleration was enabled through Compute Unified Device Architecture (CUDA) 11.8 and the CUDA Deep Neural Network library (cuDNN). The same Sionna RT backend was used throughout the scene generation and CKM synthesis pipeline.

\subsubsection{Baseline Comparison Methods}

The comparison includes four learning-based baselines: U-Net, ConvNeXt U-Net, 3D-RadioDiff, and FPN-Transformer. The baseline models are included to compare conventional encoder-decoder reconstruction, stronger convolutional feature extraction, and generative reconstruction. The diffusion baseline follows the RadioDiff approach proposed in \cite{Wang2025RadioDiff}, while the overall comparison setting is aligned with recent CKM reconstruction studies \cite{Levie2021RadioUNet,Teganya2022DeepCompletionAutoencoders,Li2024RadioGAT}. For the few-shot rapid deployment study, only FPN-Transformer is retained, since it represents the strongest deep baseline under the current framework.

\subsubsection{Training Strategy}

After height slot alignment, the input tensor contains $61$ channels in total, including $15$ static scene channels, three observation slots with $13$ channels each, and $7$ target specific channels. All deep models are trained on the seen scene training split. Unless otherwise stated, the batch size is set to $8$, the maximum training epoch number is set to $120$, the initial learning rate is set to $2 \times 10^{-4}$, and the weight decay is set to $10^{-4}$. The optimizer is AdamW, and the learning rate is updated by a cosine annealing schedule over the full training horizon. Validation is performed once every epoch, and early stopping is applied with a patience of $15$ epochs. Mixed precision training is enabled when a CUDA device is available.

The gain target is clipped to the interval from $-140\,\mathrm{dB}$ to $-30\,\mathrm{dB}$ and then linearly normalized to $[0,1]$. The input tensor is used in its original channelwise form, whereas the output channel-gain CKM slice is normalized before training and mapped back to the physical gain domain during evaluation. The valid evaluation region is defined by the flyable mask together with the no-fly constraint. For uncertainty supervision, the teacher uncertainty map is constructed from the target height risk field, the line of sight cue, and the gain span cue, with weights $0.45$, $0.30$, and $0.25$, respectively. This design is consistent with the methodology in Section~\ref{sec_method}-D, and ensures that the uncertainty head focuses on spatial regions that are both hard to predict and operationally important.

For U-Net, ConvNeXt U-Net, and FPN Transformer, the training objective is a composite masked loss. It consists of a masked $\ell_{1}$ term, a masked Charbonnier term with weight $0.6$, a masked gradient consistency term with weight $0.15$, and an uncertainty supervision term with weight $0.10$. The uncertainty term is computed only over valid target cells. For 3D-RadioDiff, the diffusion process uses $50$ forward diffusion steps, $25$ sampling steps during testing, and $12$ sampling steps during validation. The diffusion base channel number is set to $32$, the beta schedule ranges from $10^{-4}$ to $2 \times 10^{-2}$, and the diffusion loss combines noise prediction, reconstruction, and gradient terms with weights $1.0$, $0.8$, and $0.10$, respectively. 

For the few-shot adaptation setting, only the unseen scene protocol is considered. The few-shot support budget is selected from $\{0,1,5,10\}$, and the support height budget is selected from $\{1,2\}$. For FPN Transformer, the few-shot adaptation stage runs for $6$ epochs with learning rate $8 \times 10^{-5}$ and batch size $4$, while freezing the backbone and updating only the trainable adaptation part defined in the code.

\begin{table*}[t]
\centering
\caption{Overall performance comparison under different evaluation protocols.}
\label{tab:overall_protocol_comparison}
\setlength{\tabcolsep}{3pt}
\begin{tabular}{l|ccc|ccc}
\toprule
\multirow{2}{*}{Model} & \multicolumn{3}{c|}{Unseen-scene zero-shot} & \multicolumn{3}{c}{Legacy patch-random} \\
\cline{2-7}
\rule{0pt}{2.6ex}
& MAE (dB) $\downarrow$ & RMSE (dB) $\downarrow$ & PSNR (dB) $\uparrow$ & MAE (dB) $\downarrow$ & RMSE (dB) $\downarrow$ & PSNR (dB) $\uparrow$ \\
\midrule
FPN-Transformer & \textbf{1.698} & \textbf{5.347} & \textbf{26.265} & \textbf{0.268} & \textbf{1.111} & \textbf{39.911} \\
3D-RadioDiff \cite{zhao2025radiodiff} & 2.572 & 6.937 & 24.004 & 0.317 & 1.221 & 39.093 \\
U-Net & 3.859 & 8.769 & 21.969 & 0.397 & 1.320 & 38.415 \\
ConvNeXt-U-Net & 3.864 & 8.956 & 21.786 & 0.409 & 1.291 & 38.606 \\
\bottomrule
\end{tabular}
\end{table*}

\subsection{Performance Evaluation Metrics}
\label{subsec_metrics}

All evaluation metrics are computed on the valid aerial region only. Let $\mathbf{G}$ and $\hat{\mathbf{G}}$ denote the ground truth and predicted channel-gain CKM slices, respectively, and let $\mathbf{M}_{\mathrm{val}}$ denote the valid mask induced by flyability and no-fly constraints. Define
\begin{equation}
N_{\mathrm{val}}=\sum_{i,j} M_{\mathrm{val},i,j}.
\end{equation}
The main metrics, namely root mean square error (RMSE), mean absolute error (MAE), and peak signal-to-noise ratio (PSNR), are then written as
\begin{align}
\mathrm{RMSE}
&=
\sqrt{
\frac{
\sum_{i,j} M_{\mathrm{val},i,j}
\left(
\hat{G}_{i,j}-G_{i,j}
\right)^{2}
}{
N_{\mathrm{val}}
}
}, \\
\mathrm{MAE}
&=
\frac{
\sum_{i,j} M_{\mathrm{val},i,j}
\left|
\hat{G}_{i,j}-G_{i,j}
\right|
}{
N_{\mathrm{val}}
}, \\
\mathrm{PSNR}
&=
20\log_{10}
\frac{
G_{\max}-G_{\min}
}{
\mathrm{RMSE}
}.
\end{align}
Here $G_{\max}-G_{\min}$ denotes the dynamic range corresponding to the clipped target interval used during training.

The primary ranking metric in the current study is RMSE in dB, since it directly reflects the average magnitude of CKM reconstruction error in the physical domain. MAE in dB is reported together with RMSE to reflect robustness to local error spikes, and PSNR is used as an auxiliary quality indicator after denormalization. When reported in the active sensing analysis, normalized mean square error (NMSE) is additionally used to measure relative reconstruction quality. In addition to overall test metrics, the evaluation is further stratified by target height and by target height role. More specifically, the test script reports per target height results for $20\,\mathrm{m}$, $60\,\mathrm{m}$, $100\,\mathrm{m}$, and $120\,\mathrm{m}$, as well as grouped results for training height targets and extrapolation height targets. A dedicated $120\,\mathrm{m}$ focus analysis is also recorded, since this height corresponds to the main extrapolation case in the proposed benchmark.

For the generalization study, each model is evaluated under both the unseen-scene zero-shot protocol and the legacy patch-random protocol. The first protocol measures cross-scene transfer ability under held-out unseen scenes, whereas the second protocol reflects the easier same-distribution setting. For the rapid deployment study, the few-shot results are summarized as performance curves with respect to the support sample budget and the support height budget. Through this setup, the experimental evaluation can jointly assess overall accuracy, cross-height extrapolation ability, unseen scene generalization, and rapid adaptation behavior under limited support observations.

\section{Results and Analysis}
\label{sec_results}

\subsection{Overall Performance Under Different Evaluation Protocols}

Table~\ref{tab:overall_protocol_comparison} reports the overall performance under the unseen-scene zero-shot protocol and the legacy patch-random protocol. The proposed FPN-Transformer achieves the best result under both protocols, with an RMSE of $5.347\,\mathrm{dB}$ in the unseen-scene zero-shot setting and $1.111\,\mathrm{dB}$ in the legacy patch-random setting. This result confirms two important observations. First, the proposed model provides the strongest overall cross-height prediction accuracy among the compared methods. Second, the legacy patch-random protocol is substantially easier than the unseen-scene protocol, which indicates that scene-level separation is a more meaningful setting for evaluating emergency deployment capability in unfamiliar urban environments, consistent with recent discussions on dataset sufficiency and measurement-based evaluation in CKM studies \cite{Xu2024HowMuchDataCKM,Wang2025CKMAidedChannelPrediction}.

Compared with U-Net and ConvNeXt-U-Net, the proposed model shows a clear advantage in the more challenging unseen-scene setting. The performance gap is also more pronounced than that under the legacy patch-random split. This behavior is consistent with the goal of the paper, since the target scenario is not same-distribution interpolation, but rapid cross-height CKM prediction in a previously unseen scene.

\subsection{Height-Wise Cross-Height Prediction Performance}

\begin{figure}[htbp]
\centering
\includegraphics[width=\columnwidth]{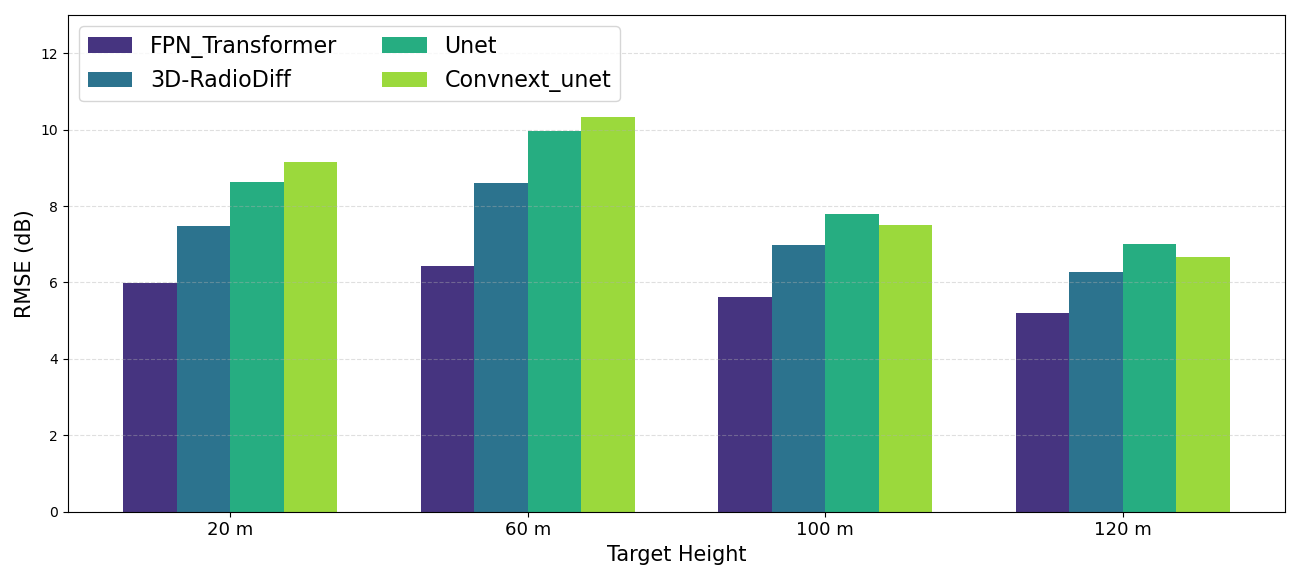}
\caption{Height-wise RMSE comparison under the unseen scene setting.}
\label{fig:heightwise_rmse_unseen}
\end{figure}

To further examine whether the model truly learns cross-height transfer rather than only achieving a favorable average score, Fig.~\ref{fig:heightwise_rmse_unseen} shows the RMSE at each target height under the unseen-scene zero-shot protocol. The proposed FPN-Transformer attains the best result at all four target heights. More specifically, its RMSE values are $5.98\,\mathrm{dB}$, $6.42\,\mathrm{dB}$, $5.62\,\mathrm{dB}$, and $5.21\,\mathrm{dB}$ at $20\,\mathrm{m}$, $60\,\mathrm{m}$, $100\,\mathrm{m}$, and $120\,\mathrm{m}$, respectively.

Several observations can be drawn from Fig.~\ref{fig:heightwise_rmse_unseen}. First, the proposed model remains consistently better than all baselines across the full target height range, which verifies that the gain is not restricted to a single layer. Second, the relative difficulty varies with target height, which indicates that cross-height prediction quality is jointly affected by altitude condition, geometric visibility, and scene specific propagation structure. Third, the proposed model also maintains the best result at the reserved $120\,\mathrm{m}$ target, which supports its extrapolation capability beyond the training height set.

\begin{figure*}[t]
\centering
\subfloat[MAE versus target height under the unseen-scene zero-shot protocol.]{
 \includegraphics[width=0.48\textwidth]{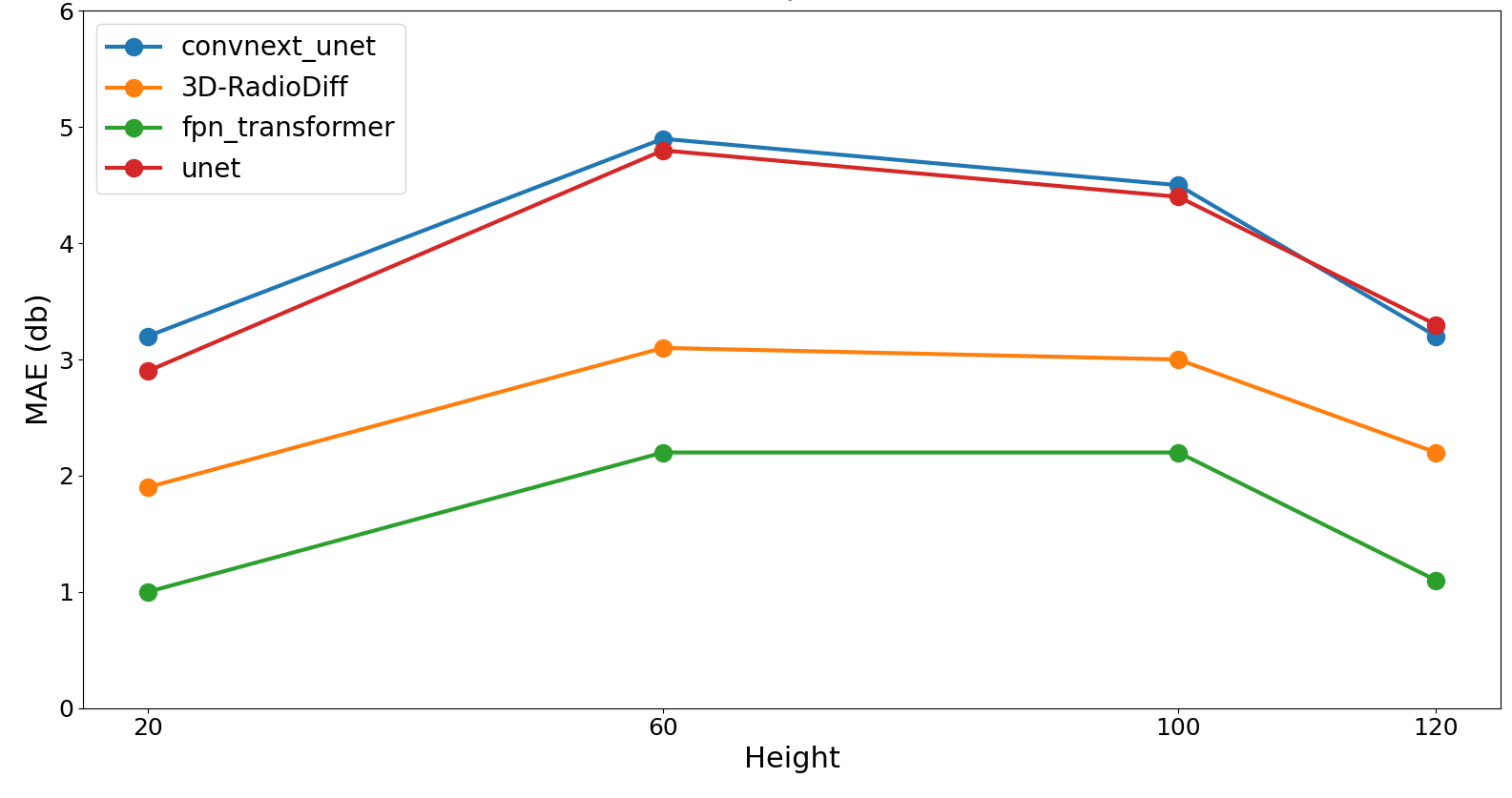}
 \label{fig:unseen_mae_height_curves}}
\hfill
\subfloat[RMSE versus target height under the unseen-scene zero-shot protocol.]{
 \includegraphics[width=0.48\textwidth]{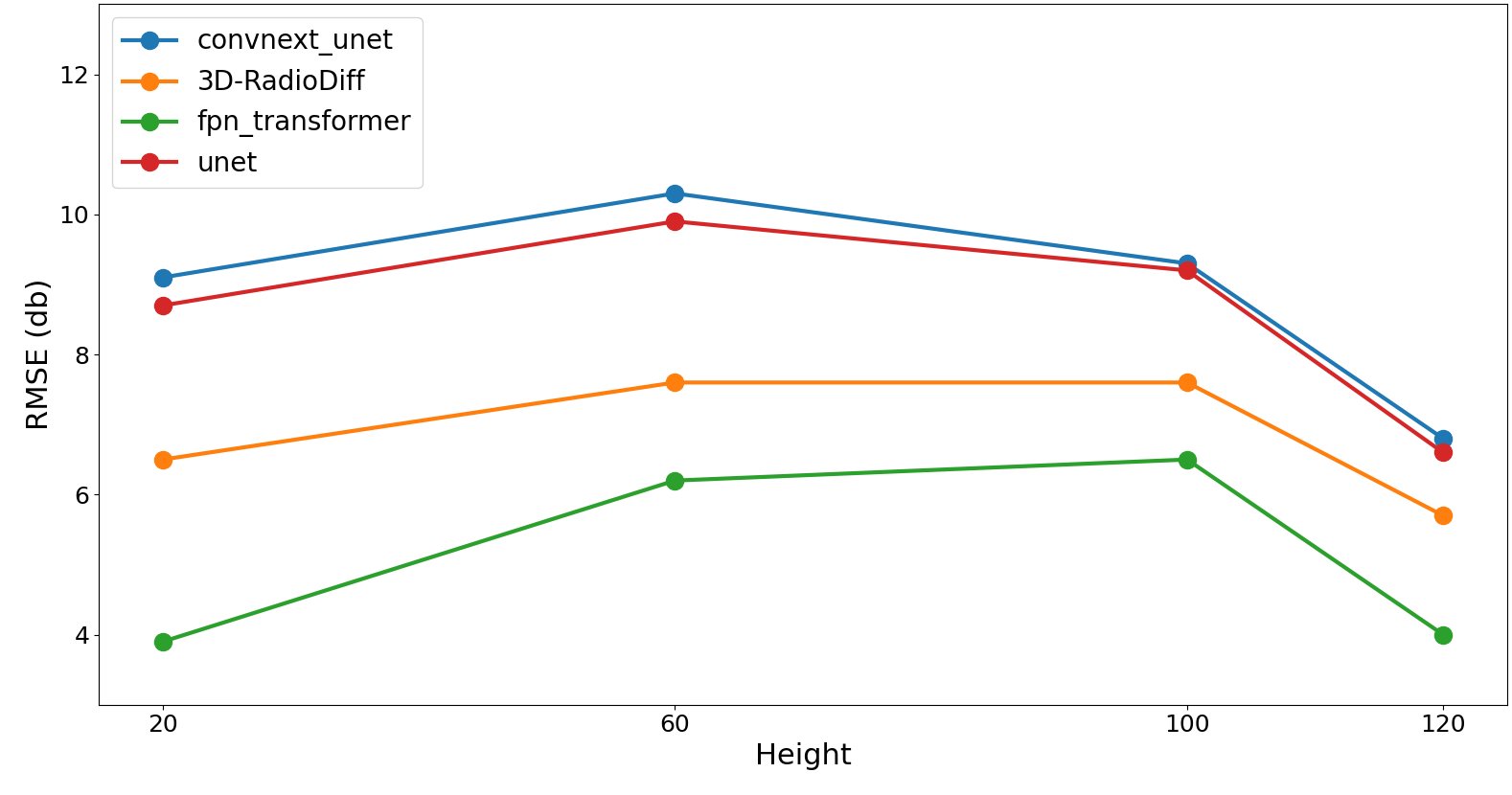}
 \label{fig:unseen_rmse_height_curves}}
\caption{Height-wise metric curves of all compared models under the unseen-scene zero-shot protocol. The proposed FPN-Transformer preserves the strongest performance across the target-height range in both MAE and RMSE.}
\label{fig:unseen_heightwise_metric_curves}
\end{figure*}

To complement the above height-wise observation, Fig.~\ref{fig:unseen_heightwise_metric_curves}(a) and Fig.~\ref{fig:unseen_heightwise_metric_curves}(b) further present the MAE and RMSE curves of all compared models under the unseen-scene zero-shot protocol. The proposed FPN-Transformer remains below the competing methods across the target-height range in both metrics. This consistency is important because it shows that the performance gain is not caused by only a few local outliers. Instead, the gain is reflected in both the absolute error level and the squared-error sensitive metric. In addition, the relative ordering between FPN-Transformer, 3D-RadioDiff \cite{zhao2025radiodiff}, U-Net, and ConvNeXt-U-Net remains largely stable across heights, which indicates that the proposed method provides a more robust cross-height representation rather than benefiting from a single favorable layer.

\begin{table}[t]
\centering
\caption{Role level RMSE under the unseen-scene zero-shot protocol. The train-height role averages the results at $20\,\mathrm{m}$, $60\,\mathrm{m}$, and $100\,\mathrm{m}$. The extrapolation role corresponds to the reserved $120\,\mathrm{m}$ target.}
\label{tab:rolewise_rmse_unseen}
\setlength{\tabcolsep}{2pt}
\begin{tabular}{lcc}
\toprule
Model & Train-height RMSE (dB) $\downarrow$ & Extrapolation RMSE (dB) $\downarrow$ \\
\midrule
FPN-Transformer & \textbf{6.01} & \textbf{5.21} \\
3D-RadioDiff \cite{zhao2025radiodiff} & 7.69 & 6.28 \\
U-Net & 8.80 & 7.01 \\
ConvNeXt-U-Net & 9.00 & 6.66 \\
\bottomrule
\end{tabular}
\end{table}

Table~\ref{tab:rolewise_rmse_unseen} summarizes the role level behavior. In this benchmark, the reserved $120\,\mathrm{m}$ target is not uniformly harder than every seen height, which suggests that cross-height extrapolation difficulty is not monotonic with altitude itself. Instead, it is closely related to how line-of-sight condition and blockage pattern evolve across scenes. Even under this more complex behavior, FPN-Transformer still preserves the strongest accuracy in both aggregated roles.

\subsection{Qualitative Visualization of Multi-Height Prediction Results}

\begin{figure}[htbp]
\centering
\subfloat[Target height $20\,\mathrm{m}$. Observed heights are $60\,\mathrm{m}$ and $100\,\mathrm{m}$.]{
\includegraphics[width=\columnwidth]{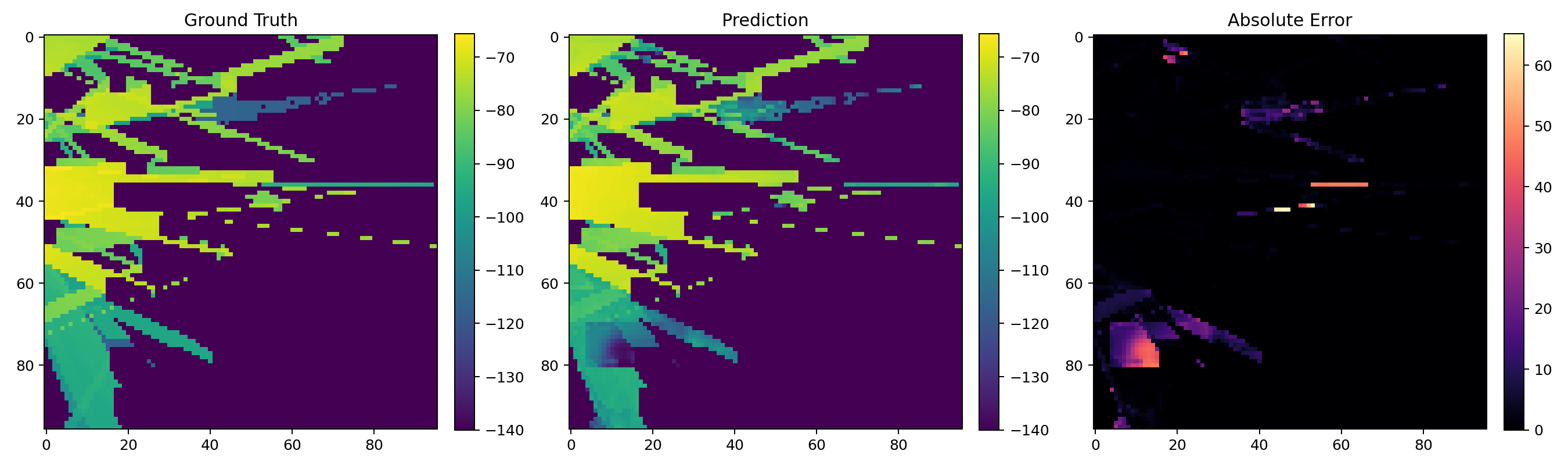}
\label{fig:qualitative_h20}}
\vspace{0.1em}

\subfloat[Target height $60\,\mathrm{m}$. Observed heights are $20\,\mathrm{m}$ and $100\,\mathrm{m}$.]{
\includegraphics[width=\columnwidth]{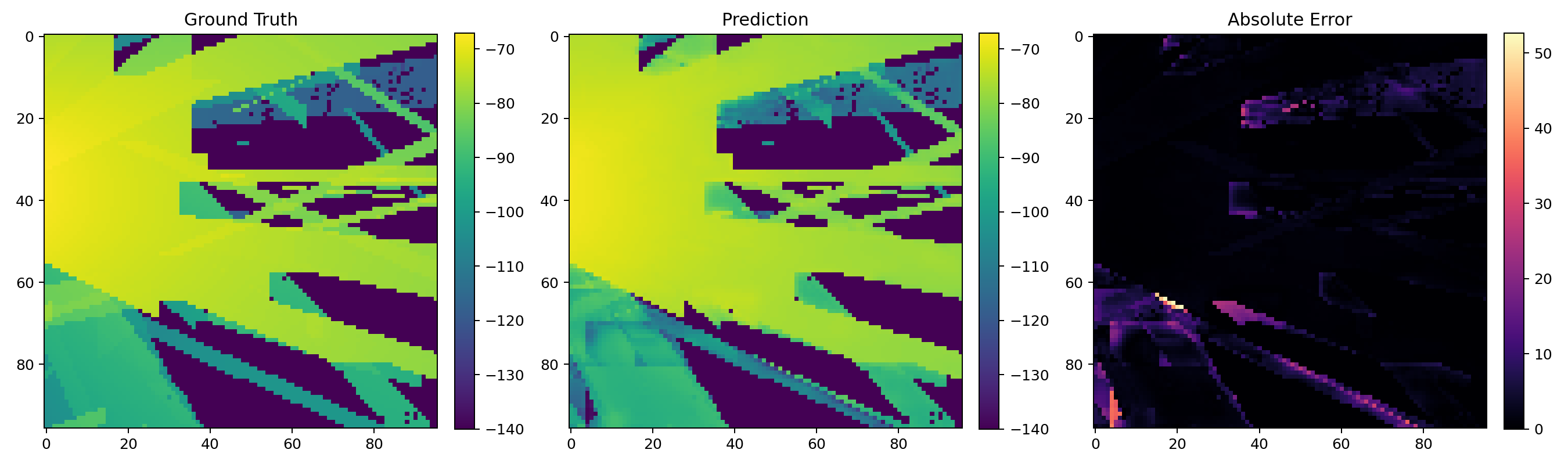}
\label{fig:qualitative_h60}}
\vspace{0.1em}

\subfloat[Target height $100\,\mathrm{m}$. Observed heights are $20\,\mathrm{m}$ and $60\,\mathrm{m}$.]{
\includegraphics[width=\columnwidth]{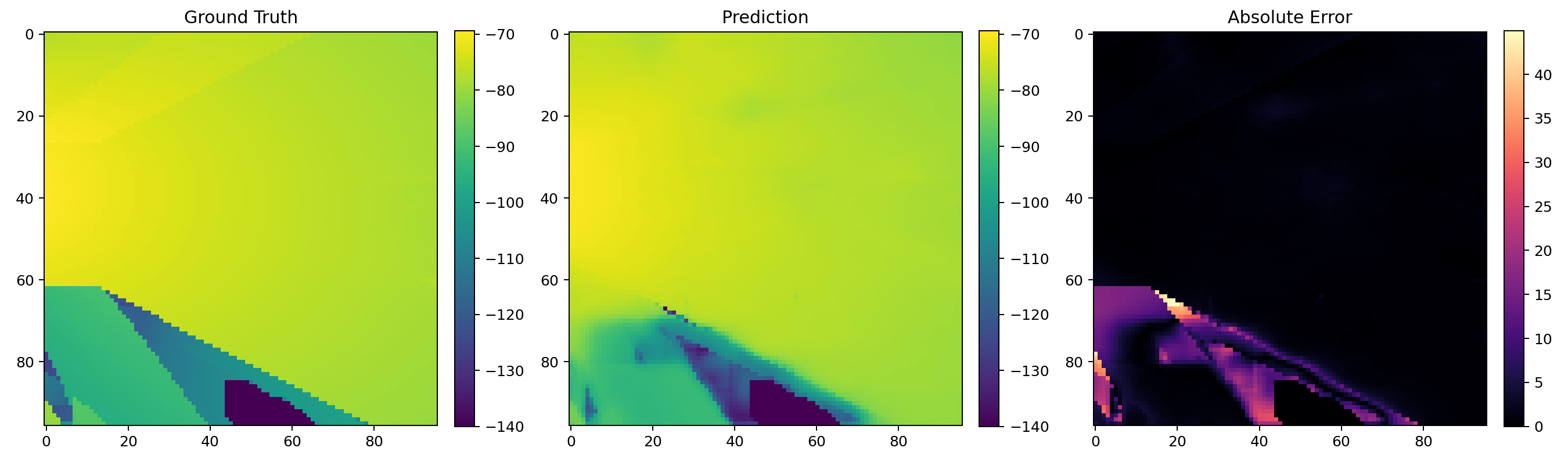}
\label{fig:qualitative_h100}}

\vspace{-0.3em}
\caption{Qualitative results of FPN-Transformer at different target heights in unseen scenes. From left to right: ground truth, prediction, and absolute error.}
\label{fig:qualitative_multiheight}
\vspace{-0.8em}
\end{figure}

Quantitative metrics indicate that the proposed model is consistently superior, but visual examples are also important for understanding what has been recovered in the spatial domain. Fig.~\ref{fig:qualitative_multiheight} presents three representative prediction cases of FPN-Transformer at target heights of $20\,\mathrm{m}$, $60\,\mathrm{m}$, and $100\,\mathrm{m}$. Each panel contains the ground truth, prediction, and absolute error from left to right.

Several qualitative patterns are evident. At $20\,\mathrm{m}$, the model accurately captures the major corridor structure and the dominant low gain region, while the remaining errors are mainly concentrated near abrupt visibility transitions and a few thin spatial structures. At $60\,\mathrm{m}$, the global pattern is also well preserved, including the large-scale coverage region and the main blocked areas, although the error becomes more noticeable around elongated transition bands. At $100\,\mathrm{m}$, the prediction still follows the dominant propagation structure, while the residual error is mainly localized near boundary-like regions and along weak coverage edges. These examples are consistent with the quantitative results and suggest that the proposed model learns both large-scale geometric regularity and height-dependent propagation variation.

\subsection{Rapid Deployment Capability Under Few Shot Adaptation}

An important goal of the proposed framework is rapid deployment in an unseen environment. Table~\ref{tab:fewshot_summary} reports the few-shot performance of FPN-Transformer under different support budgets. The zero-shot baseline starts from an RMSE of $5.347\,\mathrm{dB}$. After introducing only one support sample, the RMSE decreases to $4.982\,\mathrm{dB}$ with one support height and to $4.731\,\mathrm{dB}$ with two support heights. When the support budget increases to $5$ and $10$, the prediction error continues to decrease, and the two-height setting remains consistently better than the one-height setting.

\begin{table}[t]
\centering
\caption{Few-shot performance of FPN-Transformer on unseen scenes. Lower RMSE indicates better performance.}
\label{tab:fewshot_summary}
\setlength{\tabcolsep}{7pt}
\begin{tabular}{lcc}
\toprule
Setting & Budget & RMSE (dB) $\downarrow$ \\
\midrule
Zero shot & No support & 5.347 \\
\midrule
\multirow{3}{*}{1-height support}
& 1-shot & 4.982 \\
& 5-shot & 4.216 \\
& 10-shot & 3.962 \\
\midrule
\multirow{3}{*}{2-height support}
& 1-shot & 4.731 \\
& 5-shot & 3.874 \\
& 10-shot & 3.518 \\
\bottomrule
\end{tabular}
\end{table}

These results show that the proposed model can rapidly exploit a very small amount of scene specific support data. The improvement from zero-shot to one shot is already meaningful, while the gains from $5$ shot to $10$ shot become more moderate. This behavior suggests that the adaptation process is efficient at low support budgets and gradually approaches saturation as more local observations are introduced. In addition, the consistent gain of the two-height setting confirms that cross-height support diversity is beneficial for rapid unseen-scene adaptation.

\subsection{Active Sensing Performance Under Additional Measurement Budget}

The next experiment studies whether the uncertainty output can be converted into effective online sensing decisions. Table~\ref{tab:active_budget_rmse} reports the RMSE under different additional sensing budgets. Here, Budget $=0$ denotes the initial reconstruction under the sequential active sensing protocol before collecting any additional online measurements. Therefore, this initialization should not be interpreted as the same quantity as the overall zero-shot RMSE in Table~\ref{tab:overall_protocol_comparison}. As the sensing budget increases, all policies improve, but the uncertainty guided policies improve more rapidly than random aerial sampling and nearest greedy travel.

\begin{table}[t]
\centering
\caption{RMSE under different sensing budgets. Lower values indicate better reconstruction quality.}
\label{tab:active_budget_rmse}
\resizebox{\columnwidth}{!}{
\begin{tabular}{lccccc}
\toprule
Policy & Budget $=0$ & Budget $=5$ & Budget $=10$ & Budget $=20$ & Budget $=40$ \\
\midrule
Random aerial sampling & 6.94 & 6.53 & 6.21 & 5.97 & 5.84 \\
Nearest greedy travel & 6.94 & 6.41 & 6.02 & 5.71 & 5.52 \\
Uncertainty-only & 6.94 & 6.16 & 5.73 & 5.31 & 5.08 \\
Uncertainty + path cost & 6.94 & 6.02 & 5.54 & 5.03 & \textbf{4.79} \\
\bottomrule
\end{tabular}
}
\end{table}

The proposed uncertainty + path cost policy achieves the best result among all practical strategies at every nonzero budget, and the performance gain becomes more evident as the budget increases. This behavior shows that uncertainty alone is already informative for selecting valuable measurements, while incorporating path cost further improves the efficiency of sensing actions under limited motion resources.

\section{Conclusion}
\label{sec_conclusion}

This paper investigated height-conditioned cross-height CKM prediction for UAV-assisted communications in geometry-rich urban environments. We formulated dense target-height CKM reconstruction from sparse multi-altitude observations under aerial feasibility and safety constraints, and proposed a geometry-aware FPN-Transformer framework that fuses scene priors, height-indexed observations, and target-height descriptors, with an auxiliary uncertainty head for reliability-aware deployment. Experiments on the layered aerial CKM benchmark showed that the proposed framework achieves the best performance under both unseen-scene zero-shot and legacy patch-random protocols, remains effective across target and extrapolation heights, and can be rapidly adapted to unseen scenes with limited few-shot support. The learned uncertainty further enables cost-aware UAV sensing, improving online CKM reconstruction under limited measurement budgets. Future work will extend the framework to real measurement campaigns, multi-transmitter and multi-user scenarios, and temporally coupled UAV sensing with stronger trajectory-cost optimization.


\begin{thebibliography}{99}
\bibitem{Jin2025I2IInpaintingCKM}
Z. Jin, L. You, J. Wang, X.-G. Xia, and X. Gao, ``An I2I Inpainting Approach for Efficient Channel Knowledge Map Construction,'' \emph{IEEE Trans. Wireless Commun.}, vol. 24, no. 2, pp. 1415--1429, Feb. 2025.

\bibitem{Wang2025CKMAidedChannelPrediction}
X. Wang, Y. Shi, T. Wang, Y. Huang, Z. Hu, L. Chen, and Z. Jiang, ``Channel Knowledge Map-Aided Channel Prediction With Measurements-Based Evaluation,'' \emph{IEEE Trans. Commun.}, vol. 73, no. 5, pp. 3622--3636, May 2025.

\bibitem{Dai2025CrossAPCKM}
Z. Dai, D. Wu, X. Xu, and Y. Zeng, ``Generating CKM Using Others' Data: Cross-AP CKM Inference with Deep Learning,'' \emph{IEEE Trans. Veh. Technol.}, early access, 2025.

\bibitem{Fu2025PartialObservationDiffusionCKM}
S. Fu, Z. Wu, D. Wu, and Y. Zeng, ``Generative CKM Construction using Partially Observed Data with Diffusion Model,'' in \emph{Proc. IEEE VTC2025-Spring}, 2025.

\bibitem{Yang2025AIGCRadioMapLAE}
B. Yang, W. Zhang, and S. Zhang, ``AIGC-Based Radio Map Construction for Channel Estimation in Low-Altitude Economy,'' in \emph{Proc. IEEE/CIC ICCC Workshops}, Shanghai, China, Aug. 2025.

\bibitem{Cai2025DeepSpaceChannelEstimation}
L. Cai, G. Xu, Q. Zhang, Z. Song, and W. Zhang, ``Deep Learning Based Channel Estimation for Deep-Space Communications,'' \emph{IEEE Trans. Veh. Technol.}, vol. 74, no. 12, pp. 19743--19755, Dec. 2025.

\bibitem{Sun2026KnowledgeDrivenDL6G}
R. Sun, N. Cheng, C. Li, W. Quan, H. Zhou, Y. Wang, W. Zhang, and X. Shen, ``A Comprehensive Survey of Knowledge-Driven Deep Learning for Intelligent Wireless Network Optimization in 6G,'' \emph{IEEE Commun. Surveys Tuts.}, vol. 28, pp. 1099--1135, 2026.

\bibitem{Yang2025RadioMapBeamformingReducedPilots}
B. Yang, W. Wang, and W. Zhang, ``Radio Map-Based Beamforming Assisted With Reduced Pilots,'' \emph{IEEE Trans. Wireless Commun.}, vol. 24, no. 10, pp. 8878--8891, Oct. 2025.

\bibitem{Gao2025LEODeepSpacePerformance}
M. Gao, G. Xu, Z. Song, Q. Zhang, and W. Zhang, ``Performance Analysis of LEO Satellite-Assisted Deep Space Communication Systems,'' \emph{IEEE Trans. Aerosp. Electron. Syst.}, vol. 61, no. 5, pp. 12628--12648, Oct. 2025.

\bibitem{Wu2018AutomaticRadioMapAdaptation}
C. Wu, Z. Yang, and C. Xiao, ``Automatic Radio Map Adaptation for Indoor Localization Using Smartphones,'' \emph{IEEE Trans. Mobile Comput.}, vol. 17, no. 3, pp. 517--531, Mar. 2018.

\bibitem{Polyzos2024BayesianActiveLearningRadioMap}
K. D. Polyzos, A. Sadeghi, W. Ye, S. Sleder, K. Houssou, J. Calder, Z.-L. Zhang, and G. B. Giannakis, ``Bayesian Active Learning for Sample Efficient 5G Radio Map Reconstruction,'' \emph{IEEE Trans. Wireless Commun.}, vol. 23, no. 12, pp. 19382--19396, Dec. 2024.

\bibitem{Romero2018BlindRadioTomography}
D. Romero, D. Lee, and G. B. Giannakis, ``Blind Radio Tomography,'' \emph{IEEE Trans. Signal Process.}, vol. 66, no. 8, pp. 2055--2069, Apr. 2018.

\bibitem{Jaiswal2026TransferLearningIndoorRadioMap}
R. K. Jaiswal, M. Elnourani, S. Deshmukh, and B. Beferull-Lozano, ``A Data-Driven Transfer Learning Method for Indoor Radio Map Estimation,'' \emph{IEEE Trans. Veh. Technol.}, vol. 75, no. 3, pp. 4261--4277, Mar. 2026.

\bibitem{Martin2014AccuracyVsResolutionRTI}
R. K. Martin, A. Folkerts, and T. Heinl, ``Accuracy vs. Resolution in Radio Tomography,'' \emph{IEEE Trans. Signal Process.}, vol. 62, no. 10, pp. 2480--2490, May 2014.

\bibitem{Lee2019AdaptiveBayesianRadioTomography}
D. Lee, D. Berberidis, and G. B. Giannakis, ``Adaptive Bayesian Radio Tomography,'' \emph{IEEE Trans. Signal Process.}, vol. 67, no. 8, pp. 1964--1979, Apr. 2019.

\bibitem{Romero2024AerialBaseStationPlacementRadioMaps}
D. Romero, P. Q. Viet, and R. Shrestha, ``Aerial Base Station Placement via Propagation Radio Maps,'' \emph{IEEE Trans. Commun.}, vol. 72, no. 9, pp. 5349--5364, Sep. 2024.

\bibitem{Wu2024EnvironmentAwareHybridBeamformingCKM}
D. Wu, Y. Zeng, S. Jin, and R. Zhang, ``Environment-Aware Hybrid Beamforming by Leveraging Channel Knowledge Map,'' \emph{IEEE Trans. Wireless Commun.}, vol. 23, no. 5, pp. 4990--5005, May 2024.

\bibitem{Zhang2025SpectrumNet}
S. Zhang, S. Jiang, W. Lin, Z. Fang, K. Liu, H. Zhang, and K. Chen, ``Generative AI on SpectrumNet: An Open Benchmark of Multiband 3-D Radio Maps,'' \emph{IEEE Trans. Cogn. Commun. Netw.}, vol. 11, no. 2, pp. 886--899, Apr. 2025.

\bibitem{Chen2025Urban3DRadioMapSparse}
X. Chen, X. Zhong, Z. Zhang, L. Dai, and S. Zhou, ``High-Efficiency Urban 3D Radio Map Estimation Based on Sparse Measurements,'' \emph{IEEE Trans. Veh. Technol.}, vol. 74, no. 10, pp. 16488--16493, Oct. 2025.

\bibitem{Xu2024HowMuchDataCKM}
X. Xu and Y. Zeng, ``How Much Data Is Needed for Channel Knowledge Map Construction?,'' \emph{IEEE Trans. Wireless Commun.}, vol. 23, no. 10, pp. 13011--13021, Oct. 2024.

\bibitem{Lee2017ChannelGainCartographyLowRankSparse}
D. Lee, S.-J. Kim, and G. B. Giannakis, ``Channel Gain Cartography for Cognitive Radios Leveraging Low Rank and Sparsity,'' \emph{IEEE Trans. Wireless Commun.}, vol. 16, no. 9, pp. 5953--5966, Sep. 2017.

\bibitem{Teganya2022DeepCompletionAutoencoders}
Y. Teganya and D. Romero, ``Deep Completion Autoencoders for Radio Map Estimation,'' \emph{IEEE Trans. Wireless Commun.}, vol. 21, no. 3, pp. 1710--1724, Mar. 2022.

\bibitem{Shrestha2022DeepSpectrumCartography}
S. Shrestha, X. Fu, and M. Hong, ``Deep Spectrum Cartography: Completing Radio Map Tensors Using Learned Neural Models,'' \emph{IEEE Trans. Signal Process.}, vol. 70, pp. 1170--1185, 2022.

\bibitem{Roger2024DLRadioMapV2X}
S. Roger, M. Brambilla, B. C. Tedeschini, C. Botella-Mascarell, M. Cobos, and M. Nicoli, ``Deep-Learning-Based Radio Map Reconstruction for V2X Communications,'' \emph{IEEE Trans. Veh. Technol.}, vol. 73, no. 3, pp. 3863--3876, Mar. 2024.

\bibitem{Bazerque2010DistributedSpectrumSensingSparsity}
J. A. Bazerque and G. B. Giannakis, ``Distributed Spectrum Sensing for Cognitive Radio Networks by Exploiting Sparsity,'' \emph{IEEE Trans. Signal Process.}, vol. 58, no. 3, pp. 1847--1862, Mar. 2010.

\bibitem{Chen2025DynamicSpectrumCartography}
X. Chen, J. Wang, and Q. Huang, ``Dynamic Spectrum Cartography: Reconstructing Spatial-Spectral-Temporal Radio Frequency Map via Tensor Completion,'' \emph{IEEE Trans. Signal Process.}, vol. 73, pp. 1184--1199, 2025.

\bibitem{Liu2026DATUnetRadioMap}
K. Liu, C. Qiu, K. Chen, Q. Zheng, L. Song, and Y. Wang, ``Paying Deformable Attention to Sparse Spatial Observations for Deep Radio Map Estimation,'' \emph{IEEE Trans. Cogn. Commun. Netw.}, vol. 12, pp. 1436--1450, 2026.

\bibitem{Zhao2025RMCDDRUAV}
H. Zhao, Q. Hao, Y. He, H. Huang, H. Sari, F. Adachi, and G. Gui, ``Radio Map Reconstruction Based on Deep Denoising Regularization for UAV Communications,'' \emph{IEEE Trans. Veh. Technol.}, vol. 74, no. 6, pp. 9876--9881, Jun. 2025.

\bibitem{Li2024RadioMapPredictiveUAV}
B. Li and J. Chen, ``Radio Map-Assisted Approach for Interference-Aware Predictive UAV Communications,'' \emph{IEEE Trans. Wireless Commun.}, vol. 23, no. 11, pp. 16725--16741, Nov. 2024.

\bibitem{Wang2025RadioDiff}
X. Wang, K. Tao, N. Cheng, Z. Yin, Z. Li, Y. Zhang, and X. Shen, ``RadioDiff: An Effective Generative Diffusion Model for Sampling-Free Dynamic Radio Map Construction,'' \emph{IEEE Trans. Cogn. Commun. Netw.}, vol. 11, no. 2, pp. 738--750, Apr. 2025.

\bibitem{Mu2021IRSRobotPathPlanningRadioMap}
X. Mu, Y. Liu, L. Guo, J. Lin, and R. Schober, ``Intelligent Reflecting Surface Enhanced Indoor Robot Path Planning: A Radio Map-Based Approach,'' \emph{IEEE Trans. Wireless Commun.}, vol. 20, no. 7, pp. 4732--4747, Jul. 2021.

\bibitem{Liao2026RadioKAN}
C. Liao, X. Ge, M. He, Y. Zheng, and S. Liu, ``KAN-Based Interpretable Radio Map Prediction Framework With Symbolic Data Fusion,'' \emph{IEEE Trans. Cogn. Commun. Netw.}, vol. 12, pp. 1788--1802, 2026.

\bibitem{Kasparick2016KernelOnlineCoverageMap}
M. Kasparick, R. L. G. Cavalcante, S. Valentin, S. Stanczak, and M. Yukawa, ``Kernel-Based Adaptive Online Reconstruction of Coverage Maps With Side Information,'' \emph{IEEE Trans. Veh. Technol.}, vol. 65, no. 7, pp. 5461--5473, Jul. 2016.

\bibitem{Romero2017PSDMapQuantizedMeasurements}
D. Romero, S.-J. Kim, G. B. Giannakis, and R. Lopez-Valcarce, ``Learning Power Spectrum Maps From Quantized Power Measurements,'' \emph{IEEE Trans. Signal Process.}, vol. 65, no. 10, pp. 2547--2560, May 2017.

\bibitem{Jaiswal2026TLMoERadioMap}
R. K. Jaiswal, M. Elnourani, S. Deshmukh, and B. Beferull-Lozano, ``Leveraging Transfer Learning for Radio Map Estimation via Mixture of Experts,'' \emph{IEEE Trans. Cogn. Commun. Netw.}, vol. 12, pp. 846--863, 2026.

\bibitem{Teganya2019LocationFreeSpectrumCartography}
Y. Teganya, D. Romero, L. M. Lopez Ramos, and B. Beferull-Lozano, ``Location-Free Spectrum Cartography,'' \emph{IEEE Trans. Signal Process.}, vol. 67, no. 15, pp. 4013--4028, Aug. 2019.

\bibitem{Zeng2021SNARMUAV}
Y. Zeng, X. Xu, S. Jin, and R. Zhang, ``Simultaneous Navigation and Radio Mapping for Cellular-Connected UAV With Deep Reinforcement Learning,'' \emph{IEEE Trans. Wireless Commun.}, vol. 20, no. 7, pp. 4205--4220, Jul. 2021.

\bibitem{Sato2021SpaceFrequencyRadioMap}
K. Sato, K. Suto, K. Inage, K. Adachi, and T. Fujii, ``Space-Frequency-Interpolated Radio Map,'' \emph{IEEE Trans. Veh. Technol.}, vol. 70, no. 1, pp. 714--725, Jan. 2021.

\bibitem{Shrestha2023SpectrumSurveying}
R. Shrestha, D. Romero, and S. P. Chepuri, ``Spectrum Surveying: Active Radio Map Estimation With Autonomous UAVs,'' \emph{IEEE Trans. Wireless Commun.}, vol. 22, no. 1, pp. 627--640, Jan. 2023.

\bibitem{Romero2024TheoreticalAnalysisRME}
D. Romero, T. N. Ha, R. Shrestha, and M. Franceschetti, ``Theoretical Analysis of the Radio Map Estimation Problem,'' \emph{IEEE Trans. Wireless Commun.}, vol. 23, no. 10, pp. 13722--13737, Oct. 2024.

\bibitem{Li2024RadioGAT}
X. Li, S. Zhang, H. Li, X. Li, L. Xu, H. Xu, H. Mei, G. Zhu, N. Qi, and M. Xiao, ``RadioGAT: A Joint Model-Based and Data-Driven Framework for Multi-Band Radiomap Reconstruction via Graph Attention Networks,'' \emph{IEEE Trans. Wireless Commun.}, vol. 23, no. 11, pp. 17777--17792, Nov. 2024.

\bibitem{Zhang2024RadiomapInpaintingRestrictedAreas}
S. Zhang, T. Yu, B. Choi, F. Ouyang, and Z. Ding, ``Radiomap Inpainting for Restricted Areas Based on Propagation Priority and Depth Map,'' \emph{IEEE Trans. Wireless Commun.}, vol. 23, no. 8, pp. 9330--9344, Aug. 2024.

\bibitem{Levie2021RadioUNet}
R. Levie, C. Yapar, G. Kutyniok, and G. Caire, ``RadioUNet: Fast Radio Map Estimation With Convolutional Neural Networks,'' \emph{IEEE Trans. Wireless Commun.}, vol. 20, no. 6, pp. 4001--4015, Jun. 2021.

\bibitem{Wilson2011VarianceBasedRTI}
J. Wilson and N. Patwari, ``See-Through Walls: Motion Tracking Using Variance-Based Radio Tomography Networks,'' \emph{IEEE Trans. Mobile Comput.}, vol. 10, no. 5, pp. 612--621, May 2011.

\bibitem{Nguyen2022Survey6GIoT}
D. C. Nguyen, M. Ding, P. N. Pathirana, A. Seneviratne, J. Li, D. Niyato, O. Dobre, and H. V. Poor, ``6G Internet of Things: A Comprehensive Survey,'' \emph{IEEE Internet Things J.}, vol. 9, no. 1, pp. 359--383, Jan. 2022.


\bibitem{zeng2026urbanrtrm}
Z. Zeng, N. Wei, M. B. Mollah, K. Wang, P. L. Yeoh, F. Xu, Y. Xiu, and Z. Zhang, ``Sparse Gain Radio Map Reconstruction With Geometry Priors and Uncertainty-Guided Measurement Selection,'' \emph{arXiv preprint arXiv:2604.05788}, 2026.

\bibitem{zeng2026gackan}
Z. Zeng, K. Wang, Z. Zhang, and Y. Xiu, ``GAC-KAN: An Ultra-Lightweight GNSS Interference Classifier for GenAI-Powered Consumer Edge Devices,'' \emph{arXiv preprint arXiv:2602.11186}, 2026.

\bibitem{zeng2026phygmoe}
Z. Zeng, Y. Zhao, K. Wang, D. Niyato, Y. Xiu, L. Chen, Z. Zhang, and N. Wei, ``PhyG-MoE: A Physics-Guided Mixture-of-Experts Framework for Energy-Efficient GNSS Interference Recognition,'' \emph{arXiv preprint arXiv:2601.12798}, 2026.

\bibitem{zeng2026skanet}
Z. Zeng, Y. Zhao, K. Wang, D. Niyato, H. Shu, J. Zhao, Y. Huang, Y. Xiu, Z. Zhang, and N. Wei, ``SKANet: A Cognitive Dual-Stream Framework With Adaptive Modality Fusion for Robust Compound GNSS Interference Classification,'' \emph{arXiv preprint arXiv:2601.12791}, 2026.


\bibitem{Liu2025KAN}
Z. Liu, Y. Wang, S. Vaidya, F. Ruehle, J. Halverson, M. Solja{\v{c}}i{\'c}, T. Y. Hou, and M. Tegmark, ``KAN: Kolmogorov--Arnold Networks,'' in \emph{Proc. Int. Conf. Learn. Represent. (ICLR)}, 2025.

\bibitem{Kato2019AISAGIN}
N. Kato, Z. M. Fadlullah, F. Tang, B. Mao, S. Tani, A. Okamura, and J. Liu, ``Optimizing Space-Air-Ground Integrated Networks by Artificial Intelligence,'' \emph{IEEE Wireless Commun.}, vol. 26, no. 4, pp. 140--147, Aug. 2019.

\bibitem{Liu2018SAGINSurvey}
J. Liu, Y. Shi, Z. M. Fadlullah, and N. Kato, ``Space-Air-Ground Integrated Network: A Survey,'' \emph{IEEE Commun. Surveys Tuts.}, vol. 20, no. 4, pp. 2714--2741, 2018.

\bibitem{Saad2020Vision6G}
W. Saad, M. Bennis, and M. Chen, ``A Vision of 6G Wireless Systems: Applications, Trends, Technologies, and Open Research Problems,'' \emph{IEEE Netw.}, vol. 34, no. 3, pp. 134--142, May 2020.

\bibitem{hoydis2023sionnart}
J. Hoydis, F. A{\"i}t Aoudia, S. Cammerer, M. Nimier-David, N. Binder, G. Marcus, and A. Keller, ``Sionna RT: Differentiable Ray Tracing for Radio Propagation Modeling,'' in \emph{Proc. IEEE GLOBECOM Workshops}, pp. 317--321, 2023.

\bibitem{wang2025consistency}
Y. Wang and G. Gui, ``Consistency-guided robust learning for content-agnostic radio frequency fingerprinting,'' \emph{IEEE Commun. Lett.}, vol. 29, no. 3, pp. 610--614, Mar. 2025.

\bibitem{ai2025openset}
C. Ai, W. Sun, X. Zhang, H. Gacanin, H. Sari, F. Adachi, and G. Gui, ``Open-set automatic modulation classification using deep metric learning and OpenMax,'' in \emph{Proc. IEEE VTC2025-Spring}, 2025.

\bibitem{zeng2026urbanrtrm}
Z. Zeng, N. Wei, M. B. Mollah, K. Wang, P. L. Yeoh, F. Xu, Y. Xiu, and Z. Zhang, ``Sparse Gain Radio Map Reconstruction With Geometry Priors and Uncertainty-Guided Measurement Selection,'' \emph{arXiv preprint arXiv:2604.05788}, 2026.

\bibitem{zeng2026jsrgfnet}
Z. Zeng, H. Shu, K. Wang, L. Chen, A. Hussian, Y. Huang, J. Zhao, Y. Xiu, and Z. Zhang, ``JSR-GFNet: Jamming-to-Signal Ratio-Aware Dynamic Gating for Interference Classification in Future Cognitive Global Navigation Satellite Systems,'' \emph{arXiv preprint arXiv:2602.00042}, 2026.

\bibitem{ITURM2160}
ITU-R, ``Framework and Overall Objectives of the Future Development of IMT for 2030 and Beyond,'' International Telecommunication Union, Recommendation ITU-R M.2160-0, Nov. 2023.

\bibitem{zhao2025radiodiff}
L. Zhao, Z. Fei, X. Wang, J. Luo, and Z. Zheng, ``3D-RadioDiff: An Altitude-Conditioned Diffusion Model for 3D Radio Map Construction,'' \emph{IEEE Wireless Commun. Lett.}, vol. 14, no. 7, pp. 1969--1973, 2025.

\bibitem{Jiang2026UniRM}
X. Jiang, T. Li, Z. Xiao, K. Chen, S. Ma, Z. Wang, and K. Li, ``UniRM: A Universal Large Model for Multiband 3D Radio Map Construction,'' \emph{IEEE J. Sel. Areas Commun.}, vol. 44, 2026.

\end{thebibliography}
\end{document}